\documentclass[manuscript,screen]{acmart}
\AtBeginDocument{%
  \providecommand\BibTeX{{%
    Bib\TeX}}}

\setcopyright{acmlicensed}
\copyrightyear{2018}
\acmYear{2018}
\acmDOI{XXXXXXX.XXXXXXX}
\acmConference[Conference acronym 'XX]{Make sure to enter the correct
  conference title from your rights confirmation emai}{June 03--05,
  2018}{Woodstock, NY}
\acmISBN{978-1-4503-XXXX-X/18/06}

\usepackage{amsmath,amsfonts}
\usepackage{algorithmic} \usepackage{multirow} 
\usepackage[ruled,vlined]{algorithm2e} 
\SetAlFnt{\footnotesize}              
\SetAlgoNlRelativeSize{-1}            
\SetKwProg{Fn}{Function}{:}{end}      
\usepackage{array}
\usepackage[caption=false,font=normalsize,labelfont=sf,textfont=sf]{subfig}
\usepackage{textcomp}
\usepackage{stfloats}
\usepackage{booktabs}  
\usepackage{url}
\usepackage{verbatim}
\usepackage{graphicx}
\usepackage{colortbl}
\usepackage[table]{xcolor}
\usepackage[most]{tcolorbox}
\usepackage{wrapfig}
\usepackage{subcaption}
\usepackage{caption}
\usepackage{subfig} 
\usepackage{float}  
\definecolor{lightgreen}{rgb}{1, 0.7, 0.56} %
\usepackage{mdframed}
\usepackage{xcolor}
\begin{document}

\title{Adaptive and Robust Data Poisoning Detection and Sanitization in Wearable IoT Systems using Large Language Models}


\author{W.K.M Mithsara}
\email{malithi.mithsara@siu.edu}
\orcid{0007-2166-2176}
\affiliation{%
  \institution{Southern Illinois University}
  \city{Carbondale}
  \state{Illinois}
  \country{USA}
}

\author{Ning Yang}
\email{nyang@siu.edu}
\orcid{0003-4837-6632}
\affiliation{%
  \institution{Southern Illinois University}
  \city{Carbondale}
  \state{Illinois}
  \country{USA}
}

\author{Ahmed Imteaj}
\email{aimteaj@fau.edu}
\orcid{0002-6975-3997}
\affiliation{%
  \institution{Florida Atlantic University}
  \city{Boca Raton}
  \state{Florida}
  \country{USA}
}
\author{Hussein Zangoti}
\affiliation{%
  \institution{Jazan University}
  \city{Jizan}
  \country{Saudi Arabia}}
\email{hmzangoti@jazanu.edu.sa}
\orcid{0002-9887-339X}

\author{Abdur R. Shahid}
\email{shahid@cs.siu.edu}
\orcid{0002-3168-8907}
\affiliation{%
  \institution{Southern Illinois University}
  \city{Carbondale}
  \state{Illinois}
  \country{USA}
}

\renewcommand{\shortauthors}{Mithsara et al.}


\begin{abstract}
  The widespread integration of wearable sensing devices in Internet of Things (IoT) ecosystems, particularly in healthcare, smart homes, and industrial applications, has required robust human activity recognition (HAR) techniques to improve functionality and user experience. Although machine learning models have advanced HAR, they are increasingly susceptible to data poisoning attacks that compromise the data integrity and reliability of these systems. Conventional approaches to defending against such attacks often require extensive task-specific training with large, labeled datasets, which limits adaptability in dynamic IoT environments. This work proposes a novel framework that uses large language models (LLMs) to perform poisoning detection and sanitization in HAR systems, utilizing zero-shot, one-shot, and few-shot learning paradigms. Our approach incorporates \textit{role-play} prompting, whereby the LLM assumes the role of expert to contextualize and evaluate sensor anomalies and \textit{think step-by-step} reasoning, guiding the LLM to infer poisoning indicators in the raw sensor data and plausible clean alternatives. These strategies minimize reliance on curation of extensive datasets and enable robust, adaptable defense mechanisms in real time. We perform an extensive evaluation of the framework, quantifying detection accuracy, sanitization quality, latency, and communication cost, thus demonstrating the practicality and effectiveness of LLMs in improving the security and reliability of wearable IoT systems. 
\end{abstract}




\maketitle

\section{Introduction}
Wearable technology for human activity recognition (HAR) has become an integral component of Internet of Things (IoT) ecosystems, enabling seamless interaction between devices in applications such as fitness tracking \cite{kaewkannate2016comparison}, smart homes \cite{6740844}, and healthcare monitoring \cite{10530202, jha2024advancements}. These systems rely on interconnected sensors to recognize daily activities (e.g., walking, sitting, standing, or jogging) and provide real-time insights. To enhance accuracy for complex tasks, modern HAR systems often deploy multiple sensors placed strategically across the body \cite{ullah2024self}. Their integration in resource-constrained wearable IoT devices introduces significant challenges, including ensuring data privacy, achieving low-latency processing, and reducing reliance on network connectivity. Furthermore, the interconnected nature of IoT ecosystems raises critical concerns about data integrity and reliability~\cite{adil2024healthcare, 10.1145/3680277}. One prominent threat is data poisoning attacks, in which attackers manipulate sensor data or alter data set labels before data are integrated into IoT systems \cite{10620768, 10821483}. It is significant to detect data poisoning attacks; after detecting them, we need to sanitize the data because they compromise the performance of HAR models and the reliability and trustworthiness of the broader IoT infrastructure. In detection, we identify a data poisoning attack in the data by recognizing label flipping or feature changes, while in sanitization, we correct the attack to restore data integrity \cite{10.1145/3627536}.

Traditional methods for detecting data poisoning attacks often rely on curated datasets, making it difficult to stay updated in dynamic environments. Additionally, adapting to changes in data distribution is challenging, resulting in degraded model performance over time and a lack of flexibility in IoT data. For instance, some researchers focus on model-level detection, specifically backdoor detection using a support vector machine, which explores the intuition that data poisoning attacks alter the model's decision boundary by injecting mislabeled data points \cite{9194010}. Others, however, concentrate on data-level detection using the K-means algorithm \ cite {wang2024poisoning}. Battista et al. \cite{biggio2012poisoning} introduced the first-ever poisoning attack, a data poisoning technique targeting support vector machines (SVMs), which utilizes gradient ascent to modify training data and maximize SVM classification errors.

Additionally, some researchers propose local data training instead of centralized model training, utilizing the concept of federated learning \cite{9194010}. Federated learning aims to enhance privacy and minimize risks associated with central data storage by distributing the training process across multiple clients. However, despite its advantages, federated learning still depends on extensive data training to defend against data poisoning attacks. Furthermore, this approach may not be effective if malicious clients introduce adversarial local updates, potentially compromising and diverting the aggregated global model, this is also called a model poisoning attack \cite{10574838, korycki2023adversarial,feng2024dpfla,sun2024gan, sun2024partner}.

Recognizing these challenges, this paper contributes to the IoT domain by comprehensively exploring the strengths and limitations of leveraging large language models (LLMs) for securing Wearable IoT systems. Specifically, our research focuses on detecting and sanitizing data poisoning attacks using large language models (LLMs) that utilize zero-shot, one-shot, and few-shot learning approaches. These methods eliminate the need for additional training on large datasets, offering a more efficient and adaptable solution for identifying and mitigating data poisoning attacks. In dynamic IoT environments, LLMs have a significant advantage through their prompt engineering \cite{ji2024hargpt} techniques as they enable them to identify new attack methods without the need for retraining or large labeled datasets, thereby saving computational power and time. LLMs are uniquely positioned to detect anomalies, such as those generated by data poisoning attacks, due to their ability to process structured and unstructured data and contextual understanding of the relationships between data elements. Furthermore, in rapidly changing contexts, the flexibility of LLMs to adapt to new attacks offers a more reliable solution than traditional methods. 

Our previous works explored the application of Large Language Models (LLMs) for detecting and sanitizing data poisoning attacks in wearable IoT systems within the context of Human Activity Recognition (HAR) \cite{mithsara2024detection, mithsara2024zeroshot, Kankanamge2024_poison}. By the zero-shot integrating learning capabilities of LLMs, we developed a prompt-based framework capable of classifying sensor data, such as accelerometer and gyroscope readings, while simultaneously identifying and correcting tampered activity labels. This approach addressed significant limitations of traditional methods, such as the need for extensive labeled datasets and frequent retraining, providing a scalable and adaptive solution for real-time poisoning detection and data sanitization in dynamic environments. Furthermore, we conducted a comprehensive evaluation using models such as ChatGPT\footnote{https://openai.com/index/chatgpt/} and Gemini\footnote{https://gemini.google.com/}, demonstrating the robustness, effectiveness, and practicality of LLMs in maintaining the integrity and security of HAR systems in wearable IoT. However, these works have several limitations, including its zero-shot learning setup, which exhibits a higher level of poisoning sample detection and a lower sanitization capability.

Against this backdrop, this paper advances the use of LLMs in securing HAR by introducing a framework for detecting and correcting data poisoning attacks in wearable IoT systems. Building on prior work that employed zero-shot learning, we extend this foundation by incorporating one-shot and few-shot prompting techniques, enhancing the adaptability and robustness of LLMs in dynamic and adversarial sensing environments. Unlike existing research that typically focuses on anomaly detection or interpreting wearable sensor data using LLM, our work presents a holistic analysis of the end-to-end pipeline, evaluating poisoning detection accuracy, data sanitization quality, communication cost, response latency, and privacy leakage. Along the way, we introduce a suite of security- and deployment-focused metrics and conduct rigorous benchmarking against traditional and LLM-based baselines. This system-oriented perspective, currently missing in the literature, provides a foundation for the next generation of trustworthy, secure, and privacy-aware LLM-IoT integration in HAR systems. The contributions of this paper are highlighted as follows. 
\begin{itemize}
    \item \textbf{A Security-Oriented LLM Framework for Wearable IoT.} We propose a novel framework that uses LLMs to detect and sanitize poisoned sensor data in wearable IoT systems, shifting the focus from anomaly detection to data restoration in adversarial environments and hence, enhancing the reliability and resilience of IoT-connected devices.

    \item \textbf{Role-Playing and Step-by-Step Reasoning for Poison Detection and Sanitization.} Our framework incorporates advanced prompt engineering strategies, including role play—where the LLM assumes the role of a domain expert—and step-by-step (chain-of-thought) reasoning. These techniques enable the LLM to contextualize sensor anomalies and systematically infer both the likelihood of poisoning and plausible clean alternatives for corrupted data.

    \item \textbf{Integration of One-Shot and Few-Shot Learning for Poisoning Defense.} Extending beyond zero-shot paradigms, we incorporate one-shot and few-shot prompting techniques using GPT-3.5, GPT-4, and Gemini. This enhances the system's adaptability and robustness with minimal reliance on large, labeled datasets. 

    \item \textbf{Extensive Evaluation Across Practical Metrics.} We provide theoretical, as well as extensive experiments assessing detection accuracy, data sanitization quality, latency, and communication overhead, providing a complete view of the system's performance and deployability.

    \item \textbf{Benchmarking Against Traditional Methods.} The study benchmarks LLM-based methods against traditional supervised learning approaches, providing insights into their relative strengths, limitations, and feasibility for securing IoT systems.  
\end{itemize}

\textbf{Organization of The paper:} In the remainder of this paper, related work is presented in Section~\ref{sec:related}. Section \ref{sec:threat_model} presents the details of the threat model. The proposed framework is described in Section~\ref{sec:system}. Section~\ref{sec:evaluation} discusses the evaluation and results. Finally, Section~\ref{sec:conclusion} concludes the paper.

\section{Related Work}\label{sec:related}
In this section, we review data poisoning attacks and survey traditional defense methods for IoT sensing systems, emphasizing their critical limitations in dynamic and adversarial settings. We then introduce recent advances in large language models (LLMs) and prompt engineering, highlighting state-of-the-art LLM applications for IoT sensor data. Finally, we clarify how our proposed research advances the intersection of IoT and LLMs by addressing key gaps in robustness and adaptability under adversarial environment.

\subsection{Data Poisoning Attack and Traditional Defense Mechanisms}
These attacks involve manipulating the training dataset by an adversary to degrade the ability of a machine learning model in predicting the outcomes accurately. For instance, an adversary can add malicious data samples, enabling machine learning models to train on malicious data to affect the system's accuracy. In addition to adding malicious samples, attackers can modify data labels in specific contexts, such as human activity recognition (HAR) using sensor data. For example, Perdisci et al. \cite{perdisci2006misleading} demonstrated one of the first experiments of data poisoning in detecting computer worms in cybersecurity by targeting worm signatures. There is more research done on data poisoning attacks on federated learning for instance, Gupta et al. \cite{GUPTA2023103270} introduce a unique poisoning attack in the context of federated learning. An attacker inverts the loss function of a model by creating malicious gradients at every SGD step to create a poisoned outcome. This was evaluated using MNIST, Fashion-MNIST, and CIFAR-10 datasets. Expanding on these approaches, Sun et al. \cite{sun2021data} introduced a federated learning-based poisoning attack that calculates gradients to poison data more effectively. Li et al.\cite{li2019deep} proposes a deep reinforcement learning-based data poisoning attack in crowdsensing systems using a Partially Observable Markov Decision Process. A label-flipping poisoning attack was developed by Shahid et al. \cite{shahid2022label} on wearable sensor data in HAR and was tested on a multi-layer decision tree, perceptron, random forest, and XGBoost. They also introduce another type of poisoning attack based on context-aware spatiotemporal poisoning attacks in HAR that utilize specific patterns and conditions to alter labels effectively \cite{10620768}. The authors further extend the work for federated learning-based wearable AI~\cite{shahid2023assessing}. 

In response to these advanced poisoning techniques, several defense mechanisms have been introduced, such as data aggregation or setting the weights of parameters to help reduce the impact of the poisoned data \cite{miao2018attack} data augmentation or adding regularization to decision boundaries to prevent misclassification of data \cite{schwarzschild2021just}, and sanitization or cleaning the data before training \cite{chan2021causative}. Zheng et al. \cite{zheng2024data} introduce a bi-level optimization-based data poisoning attack and countermeasure-based defense mechanisms using Local Differential Privacy (LDP) for crowdsensing systems. Attackers achieve this by manipulating certain compromised workers to upload malicious sensory data. Consequently, defenders can identify the most damaging data by calculating corruption probabilities and formulating optimal countermeasures to detect the poisoning attack. Additionally, Feng et al. \cite{feng2024dpfla} propose another defending method in a poisoning attack in private federated learning, and they apply the removable marks to the gradient. Here, Fang et al. \cite{fan2022survey} propose defense methods such as maximizing the influence of estimation (MIE) and median of a weighted average (MWA) for enhanced resilience. Moreover, to counter label-flipping attacks, Andrea et al. \cite{paudice2019label} present a label sanitization approach using k-nearest neighbors (k-NN) to ensure consistent labels within neighborhoods.

\textbf{\textit{Limitations of Existing Research Practices:}} Existing defenses against poisoning attacks in wearable IoT systems have notable limitations, which can be summarized as follows. 
\begin{itemize}
    \item \textbf{Dependency on Curated Datasets:} Traditional methods often require meticulously curated datasets such as UCI HAR\cite{anguita2013public}, WISDM \cite{kwapisz2011activity}, PAMAP2\cite{reiss2012introducing}, which are labor-intensive to develop and maintain. Furthermore, these datasets are difficult to keep updated in dynamic, real-world environments where data evolves rapidly.

    \item \textbf{Challenges with Changing Data Distributions:} Struggle to adapt to shifts in data characteristics, leading to model drift, where model performance degrades over time. There is also need for frequent retraining to maintain accuracy, which is time-consuming and resource-intensive.

    \item \textbf{Lack of Flexibility Across Diverse IoT Data Sources:} Inability to generalize effectively across heterogeneous data streams generated by different wearable devices. Hence, methods that perform well in one scenario may fail in another due to variations in data modalities and device contexts.
\end{itemize}
It is important to note that, while these limitations are discussed in the context of poisoning attacks in wearable sensing systems, they are fundamental challenges in dynamic sensing environments more broadly. These shortcomings have motivated recent interest in exploring Large Language Models (LLMs) as a means to enhance adaptability and robustness in wearable sensing applications.

\subsection{Large Language Models for Wearable IoT}

Large language models are increasingly being applied to various tasks using sensor data in fields such as human activity recognition\cite{ji2024hargpt} and health prediction\cite{kim2024health}. Central to this progress is the practice of prompt engineering, which is the design of carefully crafted inputs that specify the desired tone, format, and scope, essential in zero-shot, one-shot, and few-shot learning to enable models to perform tasks without extensive fine-tuning\cite{sahoo2024systematic}. In zero-shot learning, prompts guide the model to generalize using only a task description, while one-shot and few-shot prompts further refine task understanding with a limited number of examples\cite{kirk2024improving, marvin2023prompt}. 

Similar to many other IoT and related domains~\cite{xiao2024efficient, boateng2025survey, 10966456, 10825187, 10697418, 10.1145/3715014.3722070}, Several recent studies exemplify these advances in wearable sensing interpretations. Ji et al.\cite{ji2024hargpt} investigated whether Large Language Models (LLMs) can perform zero-shot human activity recognition (HAR) using raw sensor data (such as sleeping, walking, bicycling, sit-stand, moving downstairs or upstairs) without prior training and using unseen data. Their proposed model, HARGPT, takes raw IMU (Inertial Measurement Unit) data as input and feeds it to LLMs with a Chain-of-thought (CoT) prompt. The HARGPT was evaluated using two datasets, Capture24 \cite{Chan2024}, containing human motion activities, and HHAR \cite{stisen2015smart}, containing downstairs and upstairs movement data with different levels of inter-class similarities. The findings demonstrate that HARGPT or LLMs can effectively interpret raw IMU data and exceed traditional classification models, such as machine learning and state-of-the-art deep learning models, with high accuracy without fine-tuning. Moreover, their findings suggest that LLMs are robust in handling unseen data samples, which can be challenging in conventional classification models, such as machine learning and deep learning.
Similar to the HARGPT paper, which examines LLMs in zero-shot human activity recognition or sensor data, the work in \cite{ferrara2024large} examines the broader potential of LLMs in wearable sensor applications, such as human activity recognition and health monitoring. Another LLM that can interpret raw sensor data is called ADL-LLM \cite{civitarese2024large}, which uses LLMs to recognize Activities of Daily Living (ADLs) using sensor data from smart homes. The ADL-LLM can effectively recognize human activities even with few or no labeled training data. Li et al. \cite{li2024sensorllm} present SENSORLLM, a novel framework that enables Large Language Models (LLMs) to perform Human Activity Recognition (HAR) from multivariate wearable sensor data using a two-stage alignment and classification approach. Here aligns sensor data is aligned with natural language using a specially created dataset (SENSORQA) of question-answer (QA) pairs. Yang et al. \cite{yang2024you} introduce LLMTrack that leverages LLMs for zero-shot trajectory tracing using real-world raw sensor data or Inertial Measurement Unit (IMU). The authors combine role-playing and step-by-step reasoning with a minimalist prompt approach to effectively predict robot movement trajectories in indoor and outdoor environments. The experiment shows that LLMTrack can surpass traditional machine learning and state-of-the-art deep learning models in accuracy and achieve over 80\% F1-score in their experiment. This approach also raised concerns related to the fact that LLMs can be used maliciously for tracking purposes without fine-tuning or in zero-shot mode. Ouyang et al. \cite{ouyang2024llmsense} present LLMSense, a system that utilizes LLMs for high-level reasoning of spatiotemporal raw sensor data. The system analyzes various activities from sensor data and applies high-level reasoning to recognize complex events such as a person suffering from a medical condition. LLMSense achieves over 80\% accuracy and demonstrates its effectiveness in edge-cloud deployment. Civitarese et al. \cite{civitarese2024large} introduce ADL-LLM, a Large Language Model designed to reason Activities of Daily Living (ADL) using sensor-based Human Activity Recognition (HAR) sensor data. The system operates in both zero-shot and few-shot settings, converting raw sensor inputs into textual representations for high-level reasoning. ADL-LLM was evaluated on real-world datasets, such as MARBLE and the UCI ADL datasets \cite{arrotta2021marble, ordonez2013activity}. In zero-shot testing, the model achieved a weighted average F1-score of 0.94 on MARBLE and 0.80 on UCI ADL datasets. Under few-shot settings, it reached up to 0.96 and 0.80 on MARBLE and UCI ADL, respectively. Hu et al. \cite{hu2025lightllm} present LightLLM, a light-based sensing pre-trained LLM that specializes in light-based sensing tasks such as indoor localization (such as office or apartment), outdoor solar-power forecasting, and indoor solar-energy estimation. The pre-trained LightLLM model can deliver more satisfactory results than state-of-the-art baselines and GPT-4. For instance, LightLLM can improve localization accuracy by up to 4.4 times and enhance indoor solar estimation by 3.4 times on unseen data.  Xu et al. \cite{xu2024autolife} propose AutoLife, a life journaling with LLMs and mobile devices. By utilizing low-cost mobile device data (from accelerometer, gyroscope, barometer, GPS, WiFi), AutoLife analyzes these sensor data and generates an automatic life journaling series of sentences, based on the user's daily tasks, such as visiting a specific place or sleeping. Next, AutoLife detects users’ contexts and fuses them to enhance precision and generate fine-grained contexts. The experiment to evaluate this approach does not utilize a real-world or public dataset as they claim there is no existing public dataset that covers life journaling, and it was conducted on four people across 58 experiments. AutoLife was evaluated using the following metrics: hallucination, precision, recall, and F1 score, achieving up to 0.0, 0.650, 0.782, and 0.704, respectively. Zhang et al. \cite{zhang2025wi} present Wi-Chat, a LLM that analyzes raw Wi-Fi signals to determine four human activity types: walking, falling, breathing, and no-event activity. Wi-Chat can analyze Channel State Information (CSI) data without traditional signal processing techniques. Wi-Chat is evaluated using a self-collected Wi-Fi CSI dataset from conventional Wi-Fi devices using two million Wi-Fi CSI packets. The model was also compared with two other baseline models (Conventional Wi-Fi-based Systems and Machine Learning Models with Raw Signals). The experiment shows that under zero-shot settings, Wi-Chat can achieve an accuracy of up to 0.90 and an F1-score of 0.90 when processing Wi-Fi signals using GPT-4o-mini with Chain-of-Thought. Fiori et al. \cite{fiori2025leveraging} assess how large language models (LLMs) can enrich Activities of Daily Living (ADLs) recognition in smart homes by introducing LLMe2e that reasons sensor-based data. LLMe2e is a zero-shot pipeline that processes raw sensor data and asks an LLM to output both the predicted Activity of Daily Living (ADL) and a plain-language rationale in a way that can avoid the cost of labeled data collection. The experiment is conducted on two publicly available datasets, the MARBLE and the UCI ADL datasets \cite{arrotta2021marble, ordonez2013activity}. The authors selected two aspects to evaluate the model: the LLMe2e recognition rate and explanation quality. The F1 weighted score for LLMe2e can reach up to 0.96 compared to another baseline model with 0.94. He et al. \cite{he2025embodiedsense} turn off-the-shelf earphones, with a stereo microphone pair and a tiny Inertial Measurement Unit (IMU), into a zero-shot, fine-grained daily activity logger in a system called EmbodiedSense.

\textbf{\textit{Common Themes in the intersection of LLM and Wearable IoT:}} Across the reviewed LLM-based HAR research, we observe several common themes towards the goal of using LLMs to understand human behavior in the physical world through wearable sensing.
\begin{itemize}
    \item \textbf{Modality Bridging:} A foundational challenge addressed in nearly all papers is the transformation of raw, multi-channel sensor data into representations that are compatible with LLMs. 

    \item \textbf{Zero/Few-Shot Inference and Cross-Dataset Generalization:} Many proposed systems leverage the generalization power of pretrained LLMs to perform activity recognition without extensive retraining. This enables zero-shot or few-shot HAR, where models can infer activities from new domains or unseen classes using language-based reasoning, while generalizing across distributional variations in datasets.


    \item \textbf{Human-Like Reasoning:} Several papers frame LLMs as expert agents capable of interpreting sensor patterns in a step-by-step, interpretable manner. This mimics the reasoning process of domain experts and enhances the system’s capacity to understand complex behavior trends.

    \item \textbf{Interpretability:} By producing textual outputs, either in the form of trend summaries, explanations, or reasoning chains, these models provide higher transparency and explainability compared to black-box deep learning models.
\end{itemize}

\subsection{Differences from the State of the Art}
While our work incorporates all of the aforementioned themes in the intersection of LLM and Wearable IoT, it diverges from current research trends by focusing on the security dimension of wearable sensing-based HAR systems using LLMs. Unlike traditional approaches that primarily detect anomalies, our work goes further by leveraging LLMs to actively restore corrupted sensor data. This is a significant shift, as in conventional anomaly detection pipelines, anomalous data is typically discarded. In contrast, our approach aims to correct corrupted data, especially valuable in attack scenarios where only a small portion of the data stream is compromised. This perspective not only improves data utility but also demonstrates the adaptive and corrective potential of LLMs in dynamic and adversarial sensing environments. 

Furthermore, unlike traditional systems built around extensive dataset curation and periodic retraining, our LLM-driven framework operates with minimal dependence on large, task-specific labeled datasets. Instead, it harnesses the zero-shot, one-shot, and few-shot generalization capabilities of LLMs to identify, explain, and repair poisoning attacks in real time. This greatly reduces reliance on costly, time-consuming, and sometimes infeasible annotation efforts. As a result, the proposed framework enables rapid deployment and continuous adaptation in environments where data distributions may shift or labeled data is scarce. Ultimately, this has broader implications for enhancing the resilience and trustworthiness of LLM-integrated HAR systems in real-world deployments.





\section{Threat Model}\label{sec:threat_model}
\begin{figure}
    \centering
    \includegraphics[width=0.85\linewidth]{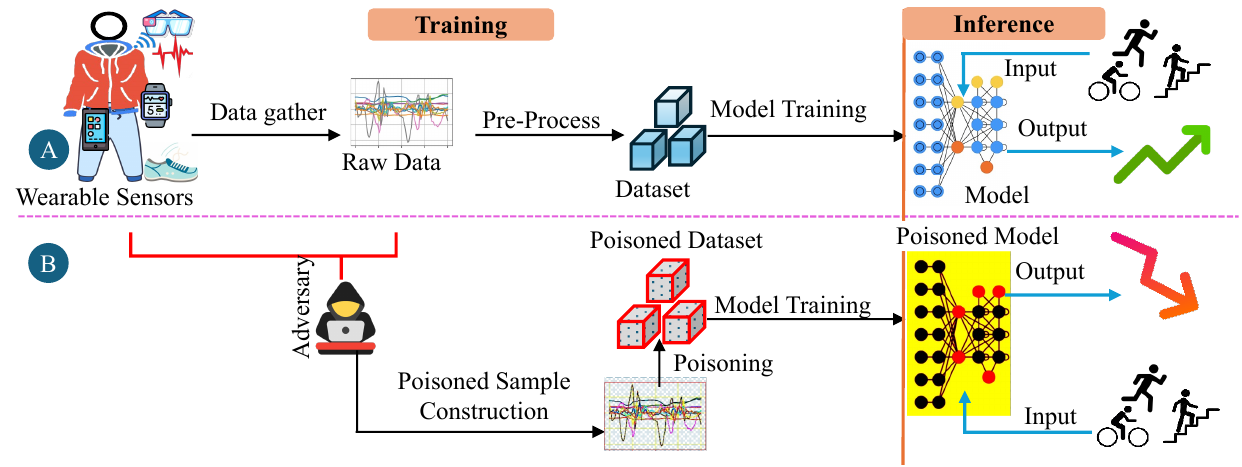}
    \Description{Illustration of the threat model showing how an adversary injects poisoned data during model training, leading to compromised performance compared to normal training flow.}
    \caption{\textbf{Overview of the Threat Model:} An adversary introduces manipulated data to poison the model, compromising its performance and accuracy. (A): Normal flow of model training, (B): Model training after an attack with poisoned data.}
    \label{fig:threat-model}
\end{figure}

Fig. \ref{fig:threat-model} illustrates the threat model, which involves poisoning the original data to manipulate the training process. This model considers two key aspects: inter-class similarities and inter-class differences. In Human Activity Recognition (HAR), certain activities exhibit inter-class similarities, such as ‘standing’ and ‘sitting’ or ‘walking’ and ‘jogging,’ where motion patterns are closely related. In contrast, activities like ‘walking’ and ‘sitting’ display inter-class differences. Building on these similarities and differences between classes, we define a threat model that assumes the presence of an adversary aiming to compromise the integrity of HAR systems. 

\textbf{Attacker's Goals.} The adversary seeks to undermine the reliability of HAR systems by inducing misclassification through label poisoning, where such degradation can have severe consequences. The adversary's goals can be articulated along two key dimensions. First, in terms of \textit{degradation of model performance}, the attacker aims to reduce the overall accuracy and reliability of the HAR model. This degradation can occur either through broad performance deterioration across multiple activity classes, reducing the system’s general reliability, or through focused attacks targeting specific activities, such as consistently misclassifying sitting as standing. Second, the goal of \textit{stealth and evasion} is paramount, as the adversary seeks to ensure that the poisoning attack remains difficult to detect during both data preprocessing and post-deployment monitoring. By exploiting inter-class similarities where activity boundaries are inherently ambiguous, the adversary can introduce manipulations that blend seamlessly with legitimate data, evading traditional anomaly detection mechanisms. 

\textbf{Attack Strategies.} To achieve the outlined goals, the adversary employs data poisoning attacks that manipulate the training data of Human Activity Recognition (HAR) systems. We categorize the adversary's attack strategies into two primary approaches. \textit{(Strategy 1: Targeted Label Flipping Attacks)} The adversary selectively flips labels between pairs of activities that exhibit high feature similarity, such as sitting vs. standing or walking vs. jogging. These activities often have overlapping motion patterns in the feature space, making the mislabeled data difficult to distinguish from legitimate samples. This focused performance degradation of the system is difficult to detect using standard anomaly detection techniques due to the semantic similarities among the closely related activities. The similarity can be measured in various ways including in the form cosine similarity, Euclidean distance, or manually defined based on domain knowledge. We adapted the domain knowledge-based similarity in the experiment.
\textit{(Strategy 2: Random Label Flipping Attack)} 
The adversary randomly selects data points from the training dataset and flips their labels to arbitrary classes, without considering semantic or feature-space similarities between activities. For instance, sitting could be mislabeled as jogging, or walking could be flipped to downstairs, despite the lack of any logical correlation between these activities. The primary objective of this attack is to maximize overall performance degradation by introducing significant label noise, thereby disrupting the model’s ability to learn consistent patterns. While this attack is generally less stealthy compared to targeted label flipping, making it more susceptible to detection, it can still be highly effective, especially true in environments   where anomaly detection mechanisms are outdated, poorly tuned, or trained on data distributions that differ from the current dataset, reducing their ability to identify the attack. The success of such an attack poses broader implications for HAR systems, as it highlights vulnerabilities that persist even without sophisticated targeting strategies.

\begin{figure}
    \centering
    \includegraphics[width=.92\linewidth]{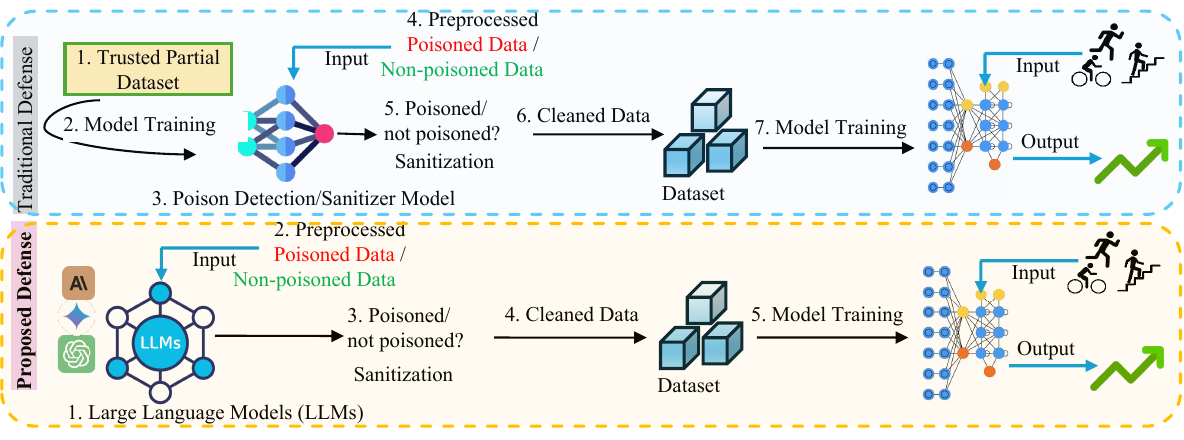}
    \Description{Diagram illustrating the LLM-driven secure wearable IoT framework. It compares traditional defenses that rely on curated trusted datasets with the proposed Large Language Model-based defense, highlighting improved adaptability and scalability in dynamic environments.}
    \caption{LLM-Driven Secure Wearable IoT: Traditional Defenses suffer from dependency on the curation of trusted datasets, limiting their adaptability and scalability in dynamic environments. We aim to address these limitations with our proposed Large Language Model (LLM)-based defense.}
    \label{fig:system-overview}
\end{figure}

\section{The Proposed LLM-based Defense Framework}\label{sec:system}

Motivated by the limitations of existing defenses, this section presents the design and methodology of the proposed LLM-driven framework for detecting and mitigating data poisoning attacks in wearable IoT systems (Fig.\ref{fig:system-overview}). The design is guided by several fundamental objectives: (i) independence from the curation of extensive trusted datasets, (ii) adaptability of evolving data distributions, and (iii) scalability across diverse environments. It is further grounded in recent advances in LLM-based reasoning, anomaly detection~\cite{xu-ding-2025-large}, and prompt-based generalization. 

Our design draws inspiration from common themes observed in emerging LLM-based HAR research. First, to address the challenge of \textbf{modalitiy bridging}, our framework converts sensor data into structured prompt templates, preserving spatiotemporal features (e.g., gravity, attitude, acceleration) while embedding them in domain-relevant natural language contexts. Second, the framework exploits the \textbf{generalization capacity} of LLMs under zero-shot and few-shot settings\cite{brown2020language, zhao2021calibrate}, which allows it to reason about poisoned activity labels, even in domains or datasets it was not explicitly trained on, without retraining or labeled supervision. Third, we incorporate \textbf{human-like reasoning} through structured prompting strategies such as role-play \cite{Chan2024} and chain-of-thought thoughts \cite{wei2022chain} to allow the LLM to perform multi-step plausibility analysis when assessing label validity. Finally, our approaches provide \textbf{interpretability} through LLMs' capability to produce natural language justification for their predictions and label sanitization, which enhances the transparency and allows for human-in-the-loop decision making when poisoned data is flagged\cite{huang2024survey}.

\begin{figure}
    \centering
    \includegraphics[width=.9\linewidth]{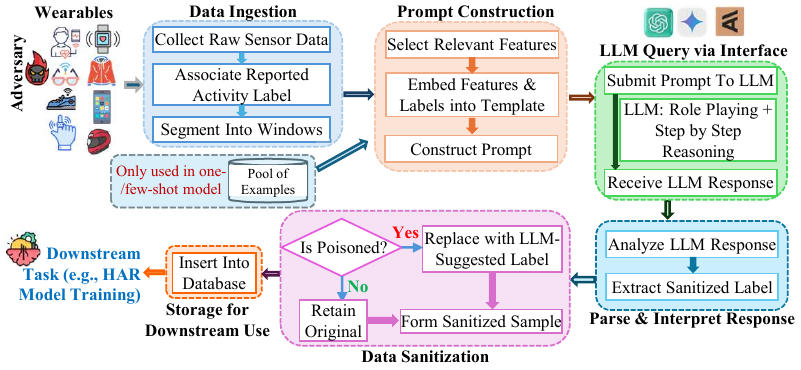}
    \Description{Flow diagram of the proposed Large Language Model-based framework for wearable IoT sensor data. It illustrates how the system detects and sanitizes data poisoning using zero-shot, one-shot, and few-shot learning, enabling interpretable reasoning and generating trustworthy datasets for robust activity recognition.}
    \caption{Overview of the proposed LLM-based framework for detecting and sanitizing data poisoning in wearable IoT sensor data. The framework enables zero-/one-/few-shot poisoning detection and sanitization with interpretable reasoning and minimal supervision for generating trustworthy datasets and improving the robustness of downstream activity recognition models.}
    \label{fig:communication}
\end{figure}

\begin{figure}
    \centering
    \footnotesize
    \setlength{\tabcolsep}{0pt}
    \begin{tabular}{ccc}
        \includegraphics[width=0.332\linewidth]{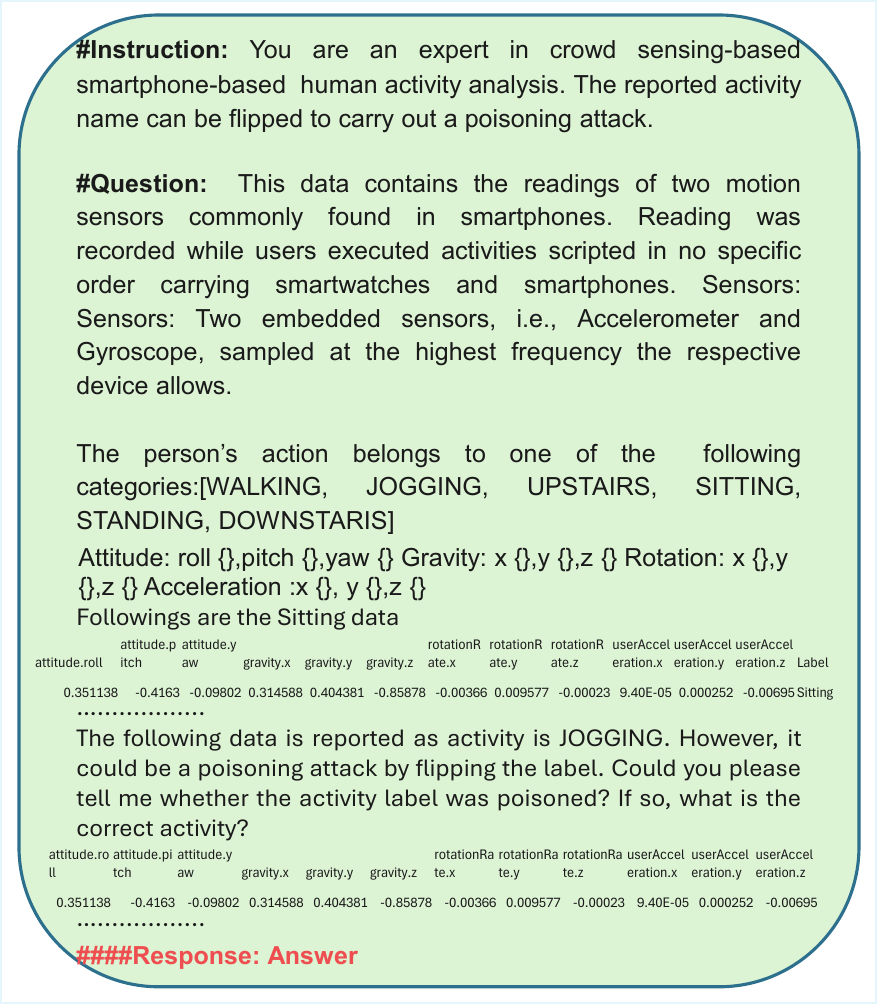} &
        \includegraphics[width=0.332\linewidth]{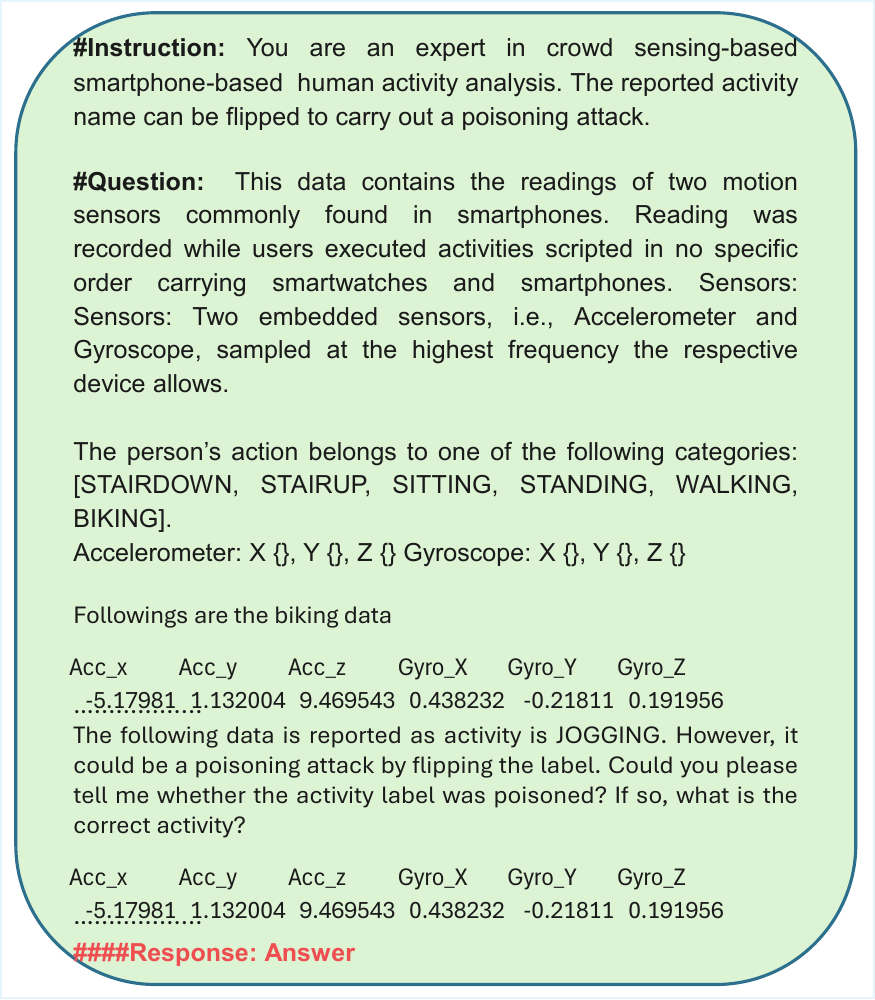} &
        \includegraphics[width=0.332\linewidth]{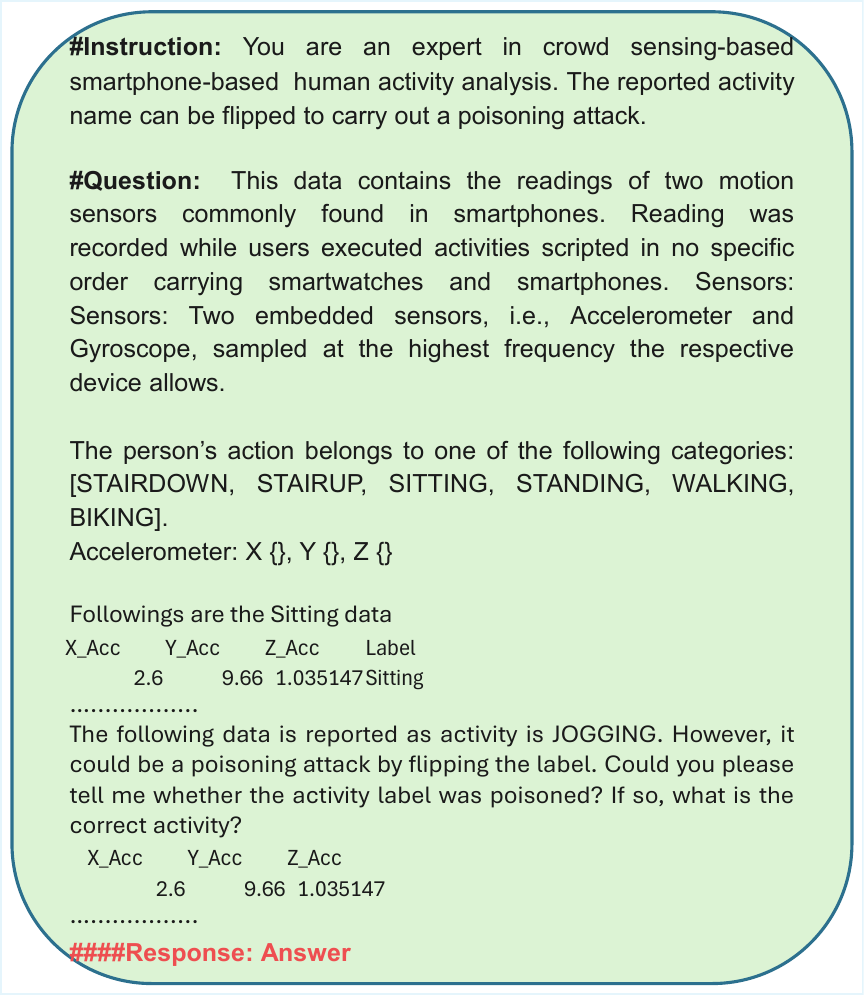} 
    \end{tabular}
    \Description{Examples of one-shot prompt templates used for Large Language Models across three datasets. The figure shows ChatGPT-3.5-turbo, ChatGPT-4.0, and Gemini prompts for MotionSense, HHAR, and WISDM datasets, respectively, illustrating how the framework generalizes across multiple sensor-text contexts.}
    \caption{One-shot prompt templates for ChatGPT-3.5-turbo, ChatGPT-4.0, and Gemini on the MotionSense (left), HHAR (middle), and WISDM (right) datasets.}
    \label{fig:oneshot-all-datasets}
\end{figure}

Building on these principles, the framework operates as a modular pipeline that integrates LLM-based detection with prompt-driven sanitization to ensure the integrity of the sensor data used for subsequent downstream tasks, such as recognition model training, as shown in Fig.\ref{fig:communication}. The process begins with the collection of raw sensor data from wearable devices, such as accelerometer and gyroscope readings, which is then preprocessed for analysis by the LLM-based defense system. Given the streaming nature of sensor data associated with physical activities, the data is processed in Windows (e.g., 100 continuous samples). Once preprocessed, the data is transmitted to a remote LLM in the form of a prompt. The prompt includes specific instructions along with a question formulated to identify potential poisoning in activity labels, utilizing zero-shot, one-shot, and few-shot prompting techniques. Fig. \ref{fig:oneshot-all-datasets} provides examples of one-shot prompt for the datasets used in the experiment. 
In the experiment, for zero-shot learning, the model is provided only with a task description and the raw sensor data to classify, without any labeled examples. For one-shot learning, we pass one example of sensor data corresponding to a specific activity, followed by new sensor data for the model to identify the activity. For few-shot learning, we provide a few labeled examples for each activity class before presenting new, unlabeled sensor data for classification. These prompts are designed to capture each dataset's unique features and apply them to real activity data with flipped labels. The prompt is then analyzed by an LLM (e.g., GPT-3.5 turbo or GPT-4, Gemini). The LLM's strength in this context stems from its zero-shot and few-shot learning abilities, which enable it to detect poisoning in previously unseen data distributions without relying on a curated, trusted dataset. During analysis, the LLM evaluates whether the data has been poisoned. If poisoning is detected, the LLM initiates a sanitization process, flagging and correcting the poisoned data. This approach is advantageous as it allows for the retention and sanitization of poisoned data, making it usable in model training rather than discarding it entirely. Following detection and sanitization by the LLMs, the framework compiles a cleaned dataset composed of sanitized and validated data. This data is then fed into the model training pipeline, ensuring that the recognition model is trained on high-integrity data, thereby enhancing the robustness and reliability of the wearable IoT system.

Algorithm \ref{alg:main_procedure} describes the detection and sanitization of data poisoning in wearable sensor data. It begins with a sequence of human activity sensor data defined as $\mathcal{D}$, a pool of labeled examples $\mathcal{E}$ for one-shot and few-shot prompting, a prompt template $\mathcal{P}$, and a remote LLM $\mathcal{L}$ to which prompts are submitted for inference. 

The process begins with data ingestion and preprocessing, where $\mathcal{D}$ is segmented into fixed-length windows ${\mathcal{D}i}{i=1}^m$ (e.g., 100 time steps per window). For each window $\mathcal{D}_i$, the algorithm selects a set of labeled examples from the pool $\mathcal{E}$, according to the chosen prompting strategy (zero-shot, one-shot, or few-shot). These examples, together with relevant sensor features from $\mathcal{D}_i$, are embedded into the prompt template $\mathcal{P}$ to generate a structured prompt $\mathcal{P}_i$. 

The constructed prompt $\mathcal{P}_i$ is then submitted to the LLM $\mathcal{L}$ via the web interface. The LLM processes the prompt and returns a natural language response $\mathcal{R}_i$, indicating whether the reported label for $\mathcal{D}_i$ is likely poisoned. If poisoning is detected, the response includes a sanitized or corrected label $\hat{y}_i$, and the corrected sample $(\mathcal{D}_i, \hat{y}_i)$ is appended to the sanitized dataset $\mathcal{D}^{\text{san}}$. If no poisoning is indicated, the original sample $(\mathcal{D}_i, y_i)$ is retained in $\mathcal{D}^{\text{san}}$. This process iterates over all windows, resulting in a sanitized dataset suitable for robust model training.

\begin{algorithm}[H]
\SetAlgoLined
\caption{LLM-Based Detection and Sanitization of Data Poisoning in Wearable Sensor Data}
\label{alg:main_procedure}
\KwIn{
    $\mathcal{D}$: Raw stream of human activity sensor data \\
    $\mathcal{E}$: Pool of labeled sensor data examples for one-shot/few-shot prompting \\
    $\mathcal{P}$: Prompt template for poisoning detection and label sanitization \\
    $\mathcal{L}$: Remote Large Language Model accessed via web interface
}
\KwOut{
    $\mathcal{D}^{\text{san}}$: Sanitized dataset for robust model training
}
\BlankLine

\textbf{Step 1: Data Ingestion and Preprocessing} \\
Segment $\mathcal{D}$ into fixed-length windows $\{ \mathcal{D}_i \}_{i=1}^{m}$\;

\textbf{Step 2: Prompt-Based Poisoning Detection and Sanitization} \\
Initialize $\mathcal{D}^{\text{san}} \leftarrow \emptyset$\;
\ForEach{window $\mathcal{D}_i$ in $\{ \mathcal{D}_i \}_{i=1}^{m}$}{
    \textbf{(a) Example Selection for Prompting} \\
    Select zero, one, or multiple labeled examples from $\mathcal{E}$ based on the prompting mode (zero-/one-/few-shot)\;

    \textbf{(b) Prompt Construction} \\
    Construct structured prompt $\mathcal{P}_i$ by embedding selected examples and features from $\mathcal{D}_i$ into the template $\mathcal{P}$\;

    Submit $\mathcal{P}_i$ to $\mathcal{L}$ and receive response $\mathcal{R}_i$\;

    \eIf{$\mathcal{R}_i$ indicates poisoning}{
        Extract sanitized label $\hat{y}_i$ or corrected data from $\mathcal{R}_i$\;
        Append $(\mathcal{D}_i, \hat{y}_i)$ to $\mathcal{D}^{\text{san}}$\;
    }{
        Append original pair $(\mathcal{D}_i, y_i)$ to $\mathcal{D}^{\text{san}}$\;
    }
}

\Return $\mathcal{D}^{\text{san}}$

\end{algorithm}

\section{Framework Analysis and Comparison}\label{sec:analysis}
We conduct a thorough analysis of the framework's performance across key areas critical to wearable IoT applications. This assessment evaluates poisoning detection accuracy, sanitization quality, latency, communication cost, and privacy leakage, providing quantitative insights into the framework's efficiency and effectiveness. Finally, we provide a detailed comparison of the proposed framework with other approaches along the key areas


\begin{table}[h]
    \centering
    \small
    \begin{tabular}{|c|c|c|c|c|c|c|}
        \hline
        \rowcolor{green!20} \multicolumn{4}{|c|}{\textbf{Sensor Data}} & \multicolumn{2}{c|}{\textbf{Poison Detection}} & \textbf{After Detecting?} \\
        \hline
        \rowcolor{pink!30} Acc\_X & Acc\_Y & Acc\_Z & \ldots &\textbf{Actual} & \textbf{Predicted} & \textbf{Label Sanitization} \\
        \hline
        \rowcolor{cyan!20} \ldots & \ldots & \ldots & \ldots & P & P &  TP \checkmark \\
        \hline
        \ldots & \ldots & \ldots & \ldots & P & N & FN\\
        \hline
        \rowcolor{cyan!20} \ldots & \ldots & \ldots & \ldots & N & P &  FP \checkmark \\
        \hline
       \ldots & \ldots & \ldots & \ldots & N & N & TN \\
         \hline
    \end{tabular}
    \caption{Illustration of sensor data poisoning detection \textbf{ (Poison as P, if not N)} and label sanitization. The table presents actual and predicted poisoning labels, followed by the corresponding sanitization action after detection. A checkmark ($\checkmark$) indicates that it goes through label sanitization after detecting poisoned data.}
    \label{tab:poison_detection}
\end{table}

\subsection{Theoretical Evaluation of The Framework}

\subsubsection{Poisoning Detection Accuracy:} Poisoning detection accuracy serves as a crucial metric for evaluating the effectiveness of our framework in accurately identifying instances of poisoned data within the continuous stream of wearable sensor inputs. This metric quantifies the system’s ability to differentiate between correctly labeled and manipulated activity labels, thereby assessing the robustness of our detection mechanism. Mathematically, poisoning detection accuracy is defined as the proportion of poisoned data instances that are correctly identified by the framework relative to the total number of data instances in the dataset. Let $ c_{i=1}^{N} $ represent the number of poisoned instances that the framework successfully detects. The total number of poisoned samples in the dataset is given by $\sum_{i=1}^{N} (x_i, y_i')$, where $ x_i $ denotes the sensor data for the $ i $-th instance, and $ y_i' $ represents the poisoned (flipped) activity label. The total number of samples both poisoned and unaltered is represented as $\sum_{i=1}^{N} \left( (x_i, y_i) \cup (x_i, y_i') \right)$, where $ y_i $ corresponds to the correct activity label for the $ i $-th sample. 
Using these definitions, the poisoning detection accuracy can be formulated as:
\begin{equation}
    \text{Detection Accuracy} = 
\frac{c_{i=1}^{N} \subseteq \left( \sum_{i=1}^{N} (x_i, y_i') \right)}
     {\sum_{i=1}^{N} \left( (x_i, y_i) \cup (x_i, y_i') \right)}
\end{equation}

In other words, we can define the poisoning detection accuracy as follows: 
\begin{equation} \label{eq:detection}
\text{Detection Accuracy} = \frac{TP + TN}{TP + TN + FP + FN} 
\end{equation}
Where, \textbf{True Positives (TP)} represents correctly identified poisoned instances, \textbf{False Negative (FN)} represents correctly identified clean (unaltered) instances, \textbf{False Positives (FP)} represents clean data wrongly classified as poisoned, and \textbf{True Negative (TN)} represents poisoned data that the detection model missed.

\subsubsection{Sanitization Quality:} Sanitization quality measures the effectiveness of the proposed framework in accurately correcting poisoned labels within a dataset. This metric is critical in evaluating how well the model distinguishes between clean and manipulated data, ensuring the integrity of the dataset used for training and inference. Unlike traditional approaches that focus solely on poisoning detection, our framework extends this by assessing the Large Language Model's (LLM) ability not only to detect poisoned data but also to correct or sanitize them effectively.

Referring to Table \ref{tab:poison_detection} and Fig. \ref{fig:communication} (right), we observe that the sanitization module processes only data that have been detected as poisoned. This means that only True Positives (TP) and False Positives (FP) reach the sanitization module, making it crucial to carefully quantify the sanitization quality based on different cases.

\textbf{Case 1: True Positives (TP) – Correctly Detected Poisoned Data.} In this scenario, the framework has successfully identified poisoned data. The sanitization quality in this case depends on how many of these TP instances are correctly sanitized. Let $CS_{TP}$ represent the number of correctly sanitized TP instances, and TP be the total number of true positives. We define the Sanitization Quality for TP instances as follows.
\begin{equation} 
SQ_{TP} = \frac{cs_{TP}}{TP} 
\end{equation}
Here, $SQ_{TP}$ represents the proportion of correctly sanitized poisoned samples. Clearly, a higher value of $SQ_{TP}$ indicates that the model is effectively removing adversarial effects while retaining legitimate patterns.

\textbf{Case 2: False Positives (FP) – Incorrectly Detected Clean Data.} In this case, the data was not poisoned but was mistakenly classified as poisoned. Here, the ideal scenario would be for the LLM to recognize the mistake and leave the sample unaltered. Therefore, the sanitization quality for FP instances is defined by how many of these false positives are left unchanged. Let $CF_{FP}$ denote the number of FP instances that the LLM did not alter (i.e., left as-is). We define the Sanitization Quality for FP instances as follows.
\begin{equation} 
SQ_{FP} = \frac{cs_{FP}}{FP} 
\end{equation}
Here, $SQ_{FP}$ represents the proportion of false positive samples that were left unchanged, preserving the original clean data. A higher value of $SQ_{FP}$  indicates that the model is successfully avoiding unnecessary modifications to clean data.

From these two metrics, we can express the overall sanitization quality as the ratio of total number of correctly processed samples and all samples reaching the sanitization module (TP+FP) as follows.
\begin{equation} 
SQ = \frac{cs_{TP} + cs_{FP}}{TP + FP} 
\end{equation}

\subsubsection{Portion of Data Remaining Poisoned:} To quantify the fact that the data is still poisoned in the data set after sanitization, that is, poisoning the data set, we must consider three cases.

\textbf{Case 1: False Negatives (FN).} These are poisoned data instances that were never detected as poisoned and therefore remain poisoned. 

\textbf{Case 2: Incorrectly Sanitized True Positives (TP).} The fraction of TP instances that were not properly sanitized $1-\frac{CS_{TP}}{TP}$, from which we get its contribution of the remaining poisoned data as $(1-\frac{CS_{TP}}{TP})\times TP$.

\textbf{Case 3: Altered False Positives (FP).} False positives (FP) are clean data mistakenly classified as poisoned. Ideally, these should be left unchanged by the sanitization module. However, if any FP instances are altered during sanitization, they effectively become poisoned. The fraction of FP instances that were modified is $1-\frac{CS_{FP}}{FP}$, from which we get its contribution of the remaining poisoned data as $(1-\frac{CS_{FP}}{FP})\times FP$.

From all these three cases, we get the total number of poisoned instances remaining in the dataset as follows.
\begin{equation}\label{eq:sanitization}
    \text{Poisoned Remaining} = FN+[(1-\frac{CS_{TP}}{TP})\times TP] + [(1-\frac{CS_{FP}}{FP})\times FP]
\end{equation}

With $(FN+TP)$ as the total poisoned data, we can express the probability of a poisoned sample from the original dataset remains poisoned in the final dataset, after applying poison detection and sanitization as follows.
\begin{equation}
    P(\textit{sample remains poisoned}) = \frac{FN+[(1-\frac{CS_{TP}}{TP})\times TP] + [(1-\frac{CS_{FP}}{FP})\times FP]}{FN+TP}
\end{equation}
A lower probability indicates a more effective defense mechanism, where most poisoned data has been successfully detected and sanitized. Conversely, a higher probability suggests that a significant amount of poisoned data persists, highlighting weaknesses in either detection or sanitization effectiveness.

\subsubsection{Communication Cost:} In the proposed framework, the communication cost primarily depends on the amount of data transferred between the data ingestion module and the LLM through the interface, as depicted in Fig. \ref{fig:communication}. Each interaction involves sending a prompt with sensor data and receiving a response on data poisoning and potential sanitization, quantified by the number of characters in both prompt and response.

Let $N$ be the total number of data samples in the sensor data stream, $W$ the fixed window size (number of samples per window), $m=\frac{N}{W}$ the total number of windows, $c_p$ the number of characters in each prompt generated from a data window, and $c_r$ the number of characters in the response from the LLM. The communication cost of sending each prompt to the LLM can be approximated by $O(c_p)$. Given that we have $m$ windows, the total cost of transmitting prompts to the LLM is $O(m.c_p) = O(\frac{N}{W}.c_p)$. After processing each prompt, the LLM returns a response of $c_r$ characters,  which provides an assessment of poisoning and, if applicable, sanitized version for the given data window. The communication cost of receving each response is therefore $O(c_r)$. With $m$ windows, the total cost of receiving response is $O(m.c_r) = O(\frac{n}{W}.c_r)$. Finally, the total communication cost, which includes both the prompt transmission and response retrieval, can be represented as follows.
\begin{equation}
    O(m.c_p + m.c_r) = O(\frac{N}{W}(c_p+c_r))
\end{equation}

\subsubsection{Response Time:} Like communication cost, latency arises from delays during LLM interactions via web interface. Total latency is influenced by data segmentation, prompt transmission, LLM processing for poisoning detection, and response retrieval. Based on our communication cost analysis, transmission and response retrieval latency are $O(m.c_p)$ and $O(m.c_r)$, respectively. With $t_p$ as the LLM's processing time per window, the total latency can be expressed as follows.
\begin{equation}
    O(m.(1+c_p+t_p+c_r)) = O(\frac{N}{W}.(1+c_p+t_p+c_r))
\end{equation}

\subsubsection{Privacy:} In our approach, privacy risks arise primarily from the communication with the Large Language Model (LLM) through the web interface, where sensitive data is transmitted for processing. With each interaction with the LLM, there is a probability $p$ that data could be intercepted. Therefore, the probability of data exposure increases with the number of interactions is as follows.
\begin{equation}
    P_{leak} = 1-(1-p)^m
\end{equation}
Thus, privacy risks grow with the number of interactions $m$, indicating a trade-off between communication frequency and data privacy in the proposed approach. 

\subsubsection{Interpretability:} A key advantage of the proposed framework is its inherent interpretability, derived from the natural language reasoning capabilities of the LLM. Unlike conventional machine learning models that operate as black boxes, the LLM generates human-readable explanations when evaluating sensor-label consistency. For each instance, the model produces a justification for its decision, explicitly stating whether the reported label aligns with the observed sensor data and explaining the rationale. For example, when analyzing a potentially poisoned activity label, the LLMs in our experiment follow an step-by-step analysis which includes (1) summarizing the expected sensor data patterns for each activity based on established domain knowledge; (2) inspecting and describing the actual sensor data values for the queried instance; (3) comparing the empirical sensor readings to the canonical patterns of all possible activities; and (4) providing a transparent, natural language explanation that makes its reasoning and conclusion explicit. These structured explanations not only support transparency and auditability of poisoning detection outcomes but also enable human analysts to verify the model's behavior.

\subsection{Comparative Analysis}
We compare our proposed API-based LLM approach with two traditional machine learning models: on-premise models and API-based cloud models. The main difference between on-premises and API-based Traditional Models is the deployment location of the model. The comparison is structured across several key criteria that highlight the strengths and limitations of each approach, as shown in the Table. \ref{tab:theory-comparison}.

Our proposed API-based LLM framework eliminates the dependency on curated, trusted datasets for poisoning detection and sanitization, making it advantageous in environments with limited labeled data. Due to LLMs ' inherent generalization capabilities, the framework also demonstrates high adaptability, handling data shifts without frequent retraining. In contrast, traditional on-premise and cloud-based models generally exhibit low adaptability, requiring periodic retraining to adjust to changing data distributions. Additionally, our approach offers flexibility across diverse IoT ecosystems, with a scalable design that accommodates various IoT contexts and device types. However, the frequent API calls required for poisoning detection and sanitization result in higher communication costs compared to on-premise models, which incur minimal communication costs by processing data locally. Regarding latency, our framework and cloud-based models experience moderate to high latency, as they depend on network speed and API response time. In contrast, on-premise models offer the lowest latency, making them ideal for time-sensitive applications. Finally, in terms of privacy, our framework presents greater risks than on-premise models due to the data sent to third-party LLM APIs, where inherent privacy risks are higher.

In terms of interpretability, our LLM-based framework provides a distinct advantage by generating natural language justifications for its poisoning assessments on the sensor data, thereby offering direct insight into the reasoning behind each decision. This stands in contrast to traditional models, which typically function as black boxes and require unreliable post hoc explanation tools to interpret their outputs.

\begin{table*}[h]
    \centering
    \captionsetup{justification=centering} 
    \caption{Comparison of our proposed LLM-based approach with different approaches.} \scriptsize
    \label{tab:theory-comparison} 
    \begin{tabular}{>{\centering}p{1.5cm}|>{\centering}p{1cm}|>{\centering}p{2cm}|>{\centering}p{2cm}|>{\centering}p{1.92cm}|>{\centering}p{1.15cm}|>{\centering}p{.9cm}|p{1.8cm}}
        \toprule
        \hline
        \rowcolor{gray!20} \textbf{Approach} & \textbf{Requires Trusted Dataset} & \textbf{Adaptability to Evolving Data Distributions} & \textbf{Flexibility in Diverse IoT Ecosystems} & \textbf{Communication Cost} & \textbf{Response Time Delay} & \textbf{Privacy} & \textbf{Interpretability} \\
        \hline
        Our Proposed LLM-based Approach 
        & \cellcolor{green!30} No 
        & \cellcolor{green!30} High; adaptable without retraining (zero-/one-/few-shot learning) 
        & \cellcolor{green!30} High; scalable across various contexts and device types 
        & \cellcolor{red!30} High (due to frequent calls to the LLM) 
        & \cellcolor{yellow!30} Moderate to High 
        & \cellcolor{red!30} Low 
        & \cellcolor{green!30} High; generates human-readable justifications \\
        \hline
        On-Premise Traditional Models \cite{schwarzschild2021just, chan2021causative, 10620768, shahid2022label} 
        & \cellcolor{red!30} Yes 
        & \cellcolor{red!30} Low; requires retraining to adapt to data shifts 
        & \cellcolor{red!30} Low; requires retraining 
        & \cellcolor{green!30} Low 
        & \cellcolor{green!30} Low 
        & \cellcolor{green!30} High 
        & \cellcolor{red!30} Low; black-box models with limited explainability \\
        \hline
        API-Based Traditional Models \cite{schwarzschild2021just, chan2021causative, 10620768, shahid2022label} 
        & \cellcolor{red!30} Yes 
        & \cellcolor{red!30} Low; requires retraining to adapt to data shifts 
        & \cellcolor{red!30} Low; requires retraining 
        & \cellcolor{red!30} High (due to frequent API calls) 
        & \cellcolor{yellow!30} Moderate to High 
        & \cellcolor{red!30} Low 
        & \cellcolor{red!30} Low; limited to post hoc interpretation tools \\
        \hline
        \multicolumn{8}{l}{*\small The main difference between On-Premise and API-based Traditional Models is the deployment location of the model.}
    \end{tabular}
\end{table*}

\subsubsection{Takeaway} While our proposed remote LLM-based framework demonstrates substantial advantages in adaptability, scalability, interpretability, and reduced dependence on curated datasets, these benefits are accompanied by trade-offs. Reliance on remote proprietary LLMs, such as GPT or Gemini, introduces communication overhead, increased response latency, and privacy risks associated with transmitting sensitive sensor data to third-party servers. 

To address these challenges, several research directions merit further exploration. These include developing techniques for compressing or summarizing sensor data before transmission, which can reduce communication overhead; investigating the on-device or edge deployment of lightweight LLMs, including open-source and small language models~\cite{10903502}, or leveraging hybrid architectures that balance local inference with cloud-based reasoning to minimize latency and communication costs; and integrating privacy-enhancing technologies, such as local differential privacy~\cite{choi2018guaranteeing}, into sensor data processing to safeguard user information. Furthermore, it is important to systematically explore the trade-offs between model complexity and resource consumption, particularly for resource-constrained edge deployments in a federated learning setting. Advancing these directions will require rigorous evaluation of robustness, adaptability, and performance of LLMs in diverse and dynamic IoT environments, ensuring reliable and efficient human activity recognition under real-world conditions.


\section{Experiment and Evaluation}\label{sec:evaluation}
\subsection{Experimental Setup}

\subsubsection{Datasets:} In this study, we employed three datasets: the MotionSense Dataset \cite{malekzadeh2018protecting}, the Heterogeneity Activity Recognition (HHAR) dataset \cite{heterogeneity_activity_recognition_344}, and Wireless Sensor Data Mining (WISDM)\cite{weiss2019WISDM}. The MotionSense dataset, collected via an iPhone 6s placed in participants' front pockets, includes six activities: walking, jogging, stairs-up, stairs-down, sitting, and standing. It captures measurements of orientation in three-dimensional space (roll, pitch, yaw), gravity, rotation, and acceleration across the x, y, and z axes. The HHAR dataset, gathered from four smartwatches and eight smartphones, records six activities—biking, sitting, standing, walking, stairsup, and stairsdown—capturing accelerometer and gyroscope data along the x, y, and z axes. The WISDM data was collected from mobile and wearable devices. The dataset includes labeled activities such as walking, jogging, sitting, standing, going upstairs, and going downstairs, captured from 36 participants. Each record contains information such as the user ID, activity label, timestamp, and accelerometer readings along the x, y, and z axes.

\subsection{Poisoning Attack Strategy}

\subsubsection{Poisoning Attack Strategy for Large Language Models (LLMs):} We randomly select continuous segments of 100 sensor data samples from the MotionSense and Heterogeneity Human Activity Recognition (HHAR) and 500 samples from the WISDM datasets to compare the dataset diversity. Each of these segments undergoes label-flipping poisoning attacks for various activities to simulate poisoning attacks, following the approach of Sijie et al. \cite{ji2024hargpt} for human activity detection via zero-shot learning. Only poisoned samples (true positives) are provided to the models throughout the experiments. Hence, our experiment does not cover the cases of false positives and true negatives. We adopted the formulas for poison detection accuracy and sanitization quality, as expressed in equations \ref{eq:detection} and \ref{eq:sanitization}, accordingly. In zero-shot learning, no poisoned example was provided. For the one-shot setting, a single example of each activity establishes a reference before introducing poisoned samples. In contrast, the few-shot setting includes a few examples per activity category to help the model recognize patterns before detecting poisoned data. Our analysis evaluates inter-class similarities and differences for label poisoning detection and sanitization across various activities, employing a one-shot prompt template (Fig. \ref{fig:oneshot-all-datasets}) in models such as ChatGPT-3.5, ChatGPT-4, and Gemini.

\subsubsection{Poisoning Attack Strategy for Traditional Methods:} Prior studies on poisoning attacks primarily address label sanitization using traditional models like k-Nearest Neighbors (k-NN) \cite{shahid2022label}, focusing more on sanitization than attack detection. This work extends the analysis by comparing conventional methods with Large Language Models (LLMs) for efficiency in sanitization, measured by time and communication costs (i.e., training sample requirements). We use k-NN, Long Short-Term Memory (LSTM) networks, and TinyBert as baselines, each trained on 40,915 samples per activity from the MotionSense, HHAR, and WISDM datasets. Testing involves 100-sample segments with poisoned labels from the MotionSense and HHAR datasets, and 500 samples from the WISDM dataset, to evaluate classification accuracy under attack.

\subsection{Evaluation Metrics}
To evaluate the efficacy of ChatGPT-3.5, ChatGPT-4, and Gemini, we employ various metrics, including detection accuracy, sanitization quality, response time or latency, and communication costs. Since our focus is on assessing the ability of LLMs to detect and sanitize poisoned data, our experiments are conducted exclusively on poisoned samples. Accordingly, we adapt the definitions of poison detection accuracy and sanitization quality to align with our experimental setup. The metrics used throughout the evaluation are as follows. (\textit{\textbf{Poison Detection Accuracy}}) This metric is defined as the ratio of correctly identified poisoned samples to the total number of poisoned samples. (\textit{\textbf{Sanitization Quality}}) Given that we evaluate only poisoned data, sanitization quality is measured as the ratio of correctly sanitized samples to the total number of correctly detected poisoned samples. (\textit{\textbf{Communication Cost}}) Communication cost is assessed by counting the number of characters in the responses generated by each LLM under zero-shot, one-shot, and few-shot scenarios. (\textit{\textbf{Response Time}}) The response time is measured using the Python `time' library, which marks the start at the beginning of processing and the end at the conclusion, followed by calculating the elapsed time in seconds. We corroborate these measurements by manually timing several samples with a stopwatch. We maintain a consistent networking environment and conduct tests at various times, including mornings around 8:00 AM, afternoons around noon, and evenings around 6:00 PM. The results show that the response times are approximately the same across different time intervals for all LLMs, with times recorded after evaluating various samples. (\textit{\textbf{Recall}}) A metric that measures a model's ability to correctly predict positive instances, in this case, poisoned data, out of all actual positive instances. We used this metric only for the traditional models. Due to our experimental setup, in the case of LLM, it is similar to poisoning detection accuracy.


\begin{figure}[h]
    \centering
    \captionsetup[subfloat]{labelformat=simple, labelsep=period}
    
    \subfloat[MotionSense Dataset]{%
        \includegraphics[width=0.31\linewidth]{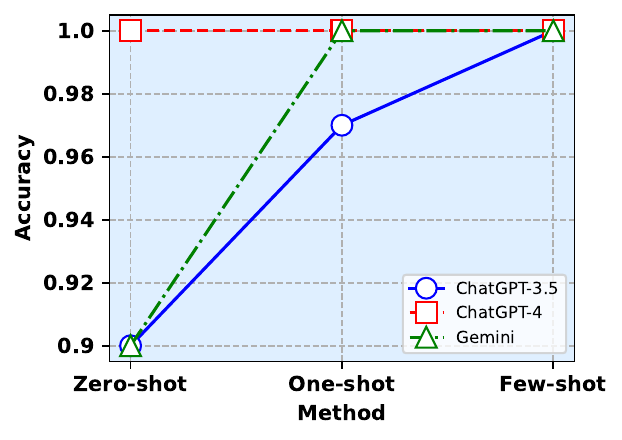}%
        \label{fig:accuracy-motion}
    }
    \hfill
    \subfloat[HHAR Dataset]{%
        \includegraphics[width=0.31\linewidth]{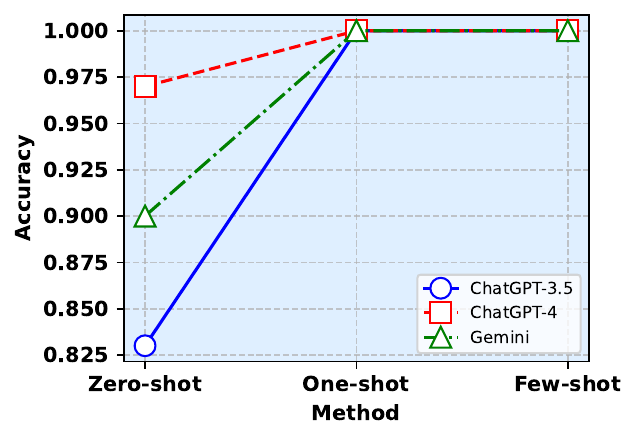}%
        \label{fig:accuracy-HHAR}
    }
    \hfill
    \subfloat[{WISDM Dataset}]{%
        \includegraphics[width=0.31\linewidth]{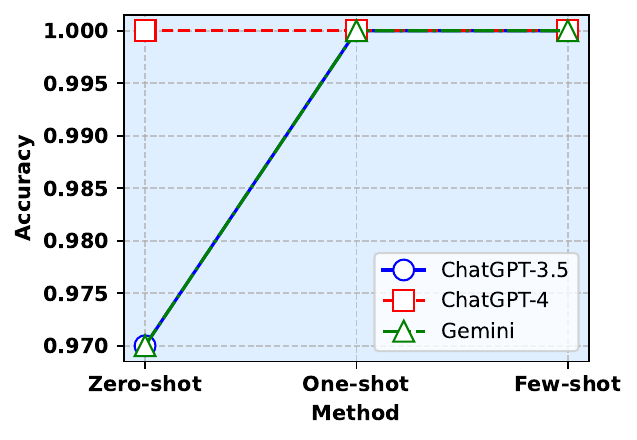}%
        \label{fig:accuracy-WISDM}
    }
    \Description{Three bar charts comparing the poison detection accuracy of zero-shot, one-shot, and few-shot methods using ChatGPT-3.5, ChatGPT-4, and Gemini across the MotionSense, HHAR, and WISDM datasets. Each subfigure shows the performance differences among the models and prompting strategies.}
    \caption{Poison Detection Accuracy Comparison of Zero-shot, One-shot, and Few-shot Methods for ChatGPT-3.5, ChatGPT-4, and Gemini on MotionSense, HHAR, and WISDM Datasets.}
    \label{fig:accuracy}
\end{figure}

\begin{figure}[h]
    \centering
    \captionsetup[subfloat]{labelformat=simple, labelsep=period}
    
    \subfloat[MotionSense Dataset]{%
        \includegraphics[width=0.31\linewidth]{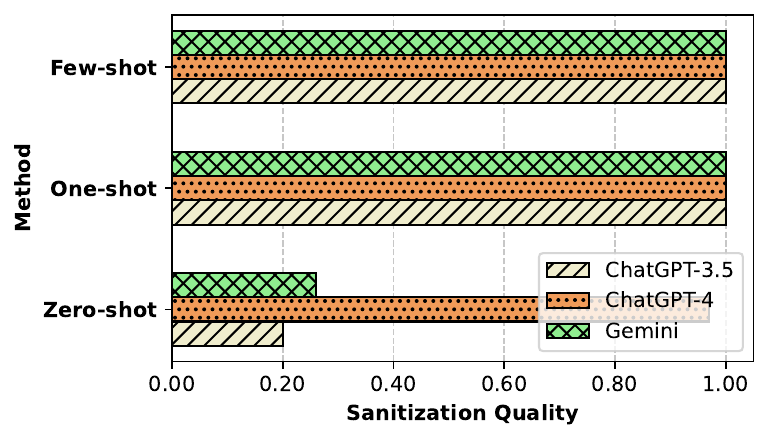}%
        \label{fig:recall-motion}
    }
    \hfill
    \subfloat[HHAR Dataset]{%
        \includegraphics[width=0.31\linewidth]{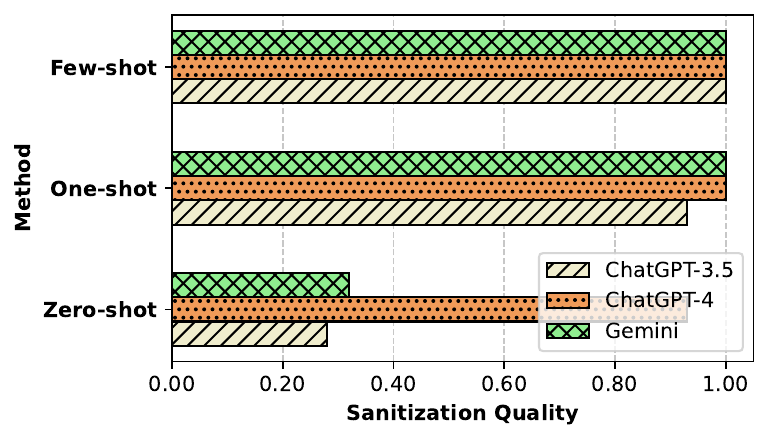}%
        \label{fig:recall-HHAR}
    }
    \hfill
    \subfloat[{WISDM Dataset}]{%
        \includegraphics[width=0.31\linewidth]{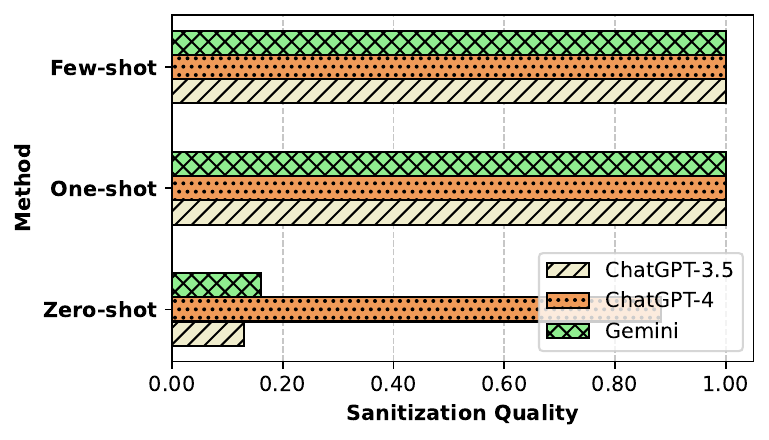}
        \label{fig:recall-WISDM}
    }

    \Description{Three bar charts comparing the sanitization quality, measured by recall, of zero-shot, one-shot, and few-shot methods using ChatGPT-3.5, ChatGPT-4, and Gemini across MotionSense, HHAR, and WISDM datasets. Each subfigure highlights how different prompting strategies influence the effectiveness of data poisoning sanitization.}
    \caption{Sanitization Quality Comparison of Zero-shot, One-shot, and Few-shot Methods for ChatGPT-3.5, ChatGPT-4, and Gemini on MotionSense, HHAR, and WISDM Datasets. }
    \label{fig:recall}
\end{figure}

\begin{figure}[h]
    \centering
    \captionsetup[subfloat]{labelformat=simple, labelsep=period}

    \subfloat[MotionSense Dataset]{%
        \includegraphics[width=0.31\linewidth]{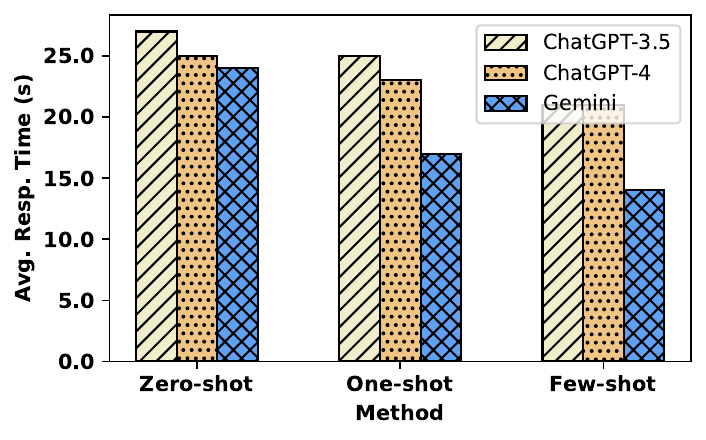}%
        \label{fig:responseTimeM}
    }
    \hspace{2mm}
    \subfloat[{HHAR Dataset}]{%
        \includegraphics[width=0.31\linewidth]{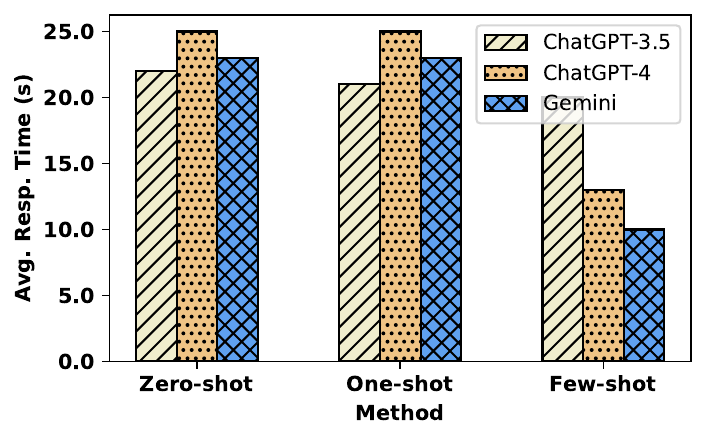}%
        \label{fig:responseTimeH}
    }
    \hspace{2mm}
    \subfloat[{{WISDM Dataset}}]{%
        \includegraphics[width=0.31\linewidth]{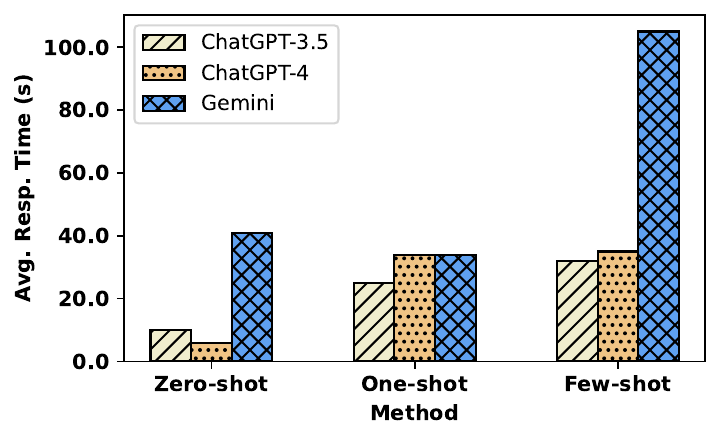}%
        \label{fig:responseTimeW}
    }

    \Description{Three bar charts showing the average response time of zero-shot, one-shot, and few-shot methods using ChatGPT-3.5, ChatGPT-4, and Gemini across MotionSense, HHAR, and WISDM datasets. The plots highlight how different prompting strategies affect the latency of poisoning detection and sanitization in wearable IoT sensor data processing.}
    \caption{Average Response Time Comparison of Zero-shot, One-shot, and Few-shot Methods for ChatGPT-3.5, ChatGPT-4, and Gemini across MotionSense, HHAR, and WISDM Datasets.}
    \label{fig:response}
\end{figure}

\begin{figure}[h]
    \centering
    \captionsetup[subfloat]{labelformat=simple, labelsep=period}
    
    \subfloat[{MotionSense Dataset}]{%
        \includegraphics[width=0.31\linewidth]{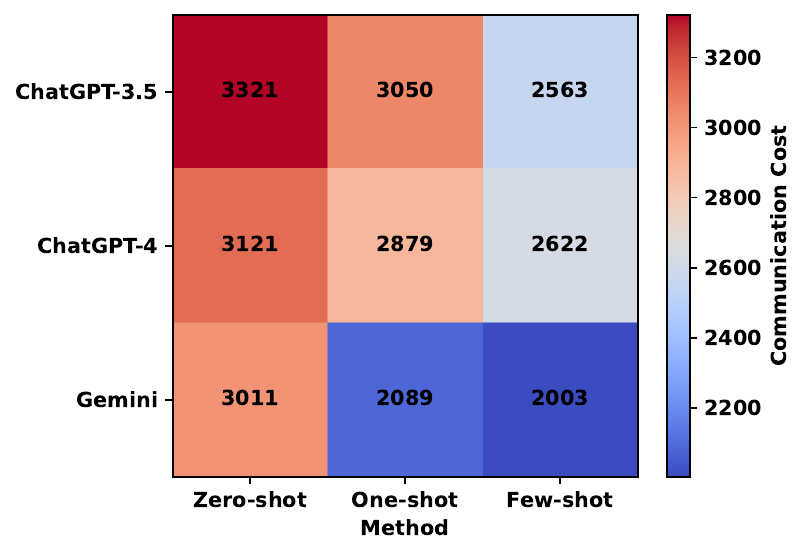}%
        \label{fig:commcostM}
    }
    \hspace{1mm}
    \subfloat[{HHAR Dataset}]{%
        \includegraphics[width=0.31\linewidth]{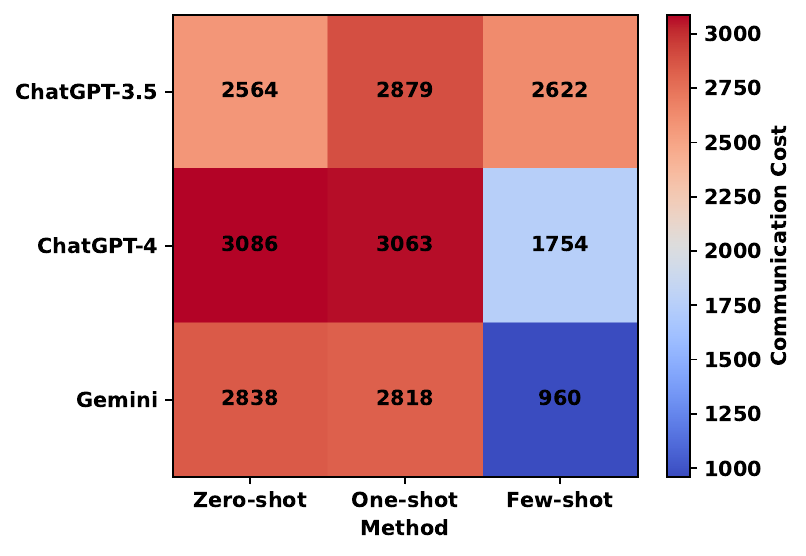}%
        \label{fig:commcostH}
    }
    \hspace{1mm}
    \subfloat[{{WISDM Dataset}}]{%
        \includegraphics[width=0.31\linewidth]{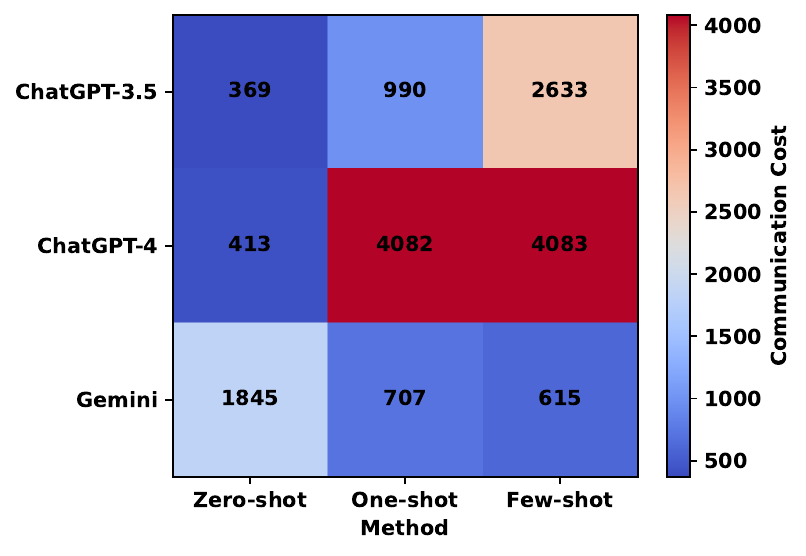}%
        \label{fig:commcostW}
    }
    \Description{Three bar charts comparing the communication cost of zero-shot, one-shot, and few-shot methods using ChatGPT-3.5, ChatGPT-4, and Gemini on the MotionSense, HHAR, and WISDM datasets. The figure shows how different prompting strategies impact the amount of data exchanged during poisoning detection and sanitization in wearable IoT systems.}
    \caption{Communication Cost Comparison of Zero-shot, One-shot, and Few-shot Methods for ChatGPT-3.5, ChatGPT-4, and Gemini on MotionSense, HHAR, and WISDM Datasets.}
    \label{fig:communicationcost}
\end{figure}


\subsection{Experiment with LLMs}

We evaluate ChatGPT-3.5, ChatGPT-4, and Gemini in their capability of detecting poisoning samples and sanitizing them in zero-shot, one-shot, and few-shot learning settings. Fig. \ref{fig:accuracy},  \ref{fig:recall}, \ref{fig:response}, and \ref{fig:communicationcost} present the accuracy, sanitization quality, response time, and communication cost among the LLMs. 

\subsubsection{Zero-shot setting:}
We provided 100 consecutive, randomly selected poisoned samples from the MotionSense and HHAR datasets for each activity in the zero-shot setting. Additionally, we conducted an experiment using 500 data points from the WISDM dataset. The final results of this experiment are presented in Fig. \ref{fig:accuracy}, \ref{fig:recall}, \ref{fig:response}, and \ref{fig:communicationcost}. This experiment builds upon our previous works \cite{mithsara2024zeroshot, mithsara2024detection, Kankanamge2024_poison}, where we focused exclusively on zero-shot learning based on inter-class similarities and differences for all datasets. Tables \ref{tab:comparison-inter-class-similarities-zero-shot-motionsense}, \ref{tab:comparison-motionsense-difference-zero}, \ref{tab:comparison-inter-class-similarities-zero-shot-hhar}, \ref{tab:comparison-interclass-difference-hhar-zero}, \ref{tab:detection_sanitization_interclass_similaraties_WISDM}, \ref{tab:detection_sanitization_difference_WISDM} present the results on zero-shot learning. In this study, we compared the accuracy and sanitization quality across each LLM to evaluate how effectively they identify data poisoning attacks using zero-shot learning. As indicated in the given tables, ChatGPT-3.5 produced incorrect results most of the time. So we expanded our study to include one-shot and few-shot learning settings to assess further the performance of Large Language Models (LLMs), which helps to evaluate improvements in the LLMs' abilities to detect and sanitize data poisoning attacks, enabling us to compare how LLMs perform when given examples rather than relying solely on zero-shot learning. As shown in Fig. \ref{fig:accuracy} and \ref{fig:recall}, ChatGPT-4 demonstrates higher accuracy and Sanitization Quality than both ChatGPT-3.5 and Gemini across the two datasets when considering overall performance in the zero-shot setting. ChatGPT-3.5 incurs a lower communication cost for the HHAR dataset and WISDM dataset as in Fig.\ref{fig:commcostH}, and \ref{fig:commcostM}, although it shows a higher cost for the MotionSense dataset(Fig. \ref{fig:commcostM}). ChatGPT-4 displays a consistent communication cost across MotionSense and HHAR datasets, but a lower communication cost for the WISDM dataset, while Gemini shows approximately similar costs. The response times follow the same pattern as the communication costs, as shown in Fig. \ref{fig:responseTimeM} and Fig. \ref{fig:responseTimeH}, depending on the volume of data processed in each response. However, for the WISDM dataset it shows a lower response time across all models. All experiments were conducted in the same lab environment, and the response times reported here represent average and approximate values.

\subsubsection{One-shot setting:}
In one-shot learning, a single sample is provided to familiarize the model with the activity before asking it to detect a poisoning attack. As in Fig. \ref{fig:accuracy} and Fig. \ref{fig:recall}, we observe that ChatGPT-3.5 achieves an accuracy of 1 and a Sanitization Quality of 0.97, demonstrating its effectiveness in detecting poisoning attacks within the MotionSense dataset. However, it sometimes fails to identify a poisoning attack, particularly when the actual label is `Standing' and the poisoned label is `Sitting', as shown in Table \ref{tab:comparison_similar}. In the HHAR dataset, ChatGPT-3.5 struggles to identify the `Stairsup' activity when the poisoned label is `Standing', suggesting a correct label of `Stairsup or Stairsdown'. Additionally, when the actual label is `Sitting' and the poisoned label is `Stairsup', it suggests `Sitting or Standing' as the correct label, as in Table \ref{tab:comparison_Difference_HHAR}. But in the WISDM dataset, ChatGPT-3.5 can correctly classify all the labels. In contrast, both ChatGPT-4 and Gemini achieve perfect accuracy and Sanitization Quality scores of 1.0 across both datasets, as in Fig. \ref{fig:recall}. When comparing communication costs and response times across all LLMs as in Fig. \ref{fig:response} and Fig. \ref{fig:communicationcost}, Chat-GPT-3.5 shows a higher communication cost than the other models for the MotionSense dataset but a lower communication cost for the HHAR dataset and WISDM. ChatGPT-4 exhibits a higher communication cost for all datasets, while Gemini shows a lower communication cost for WISDM and a higher cost for the HHAR dataset, as indicated in Fig. \ref{fig:communicationcost}. It is worth noting that we did not provide any specific instructions to the LLMs regarding the control of response length or the number of words generated for each prompt. We plan to explore this aspect in future work.

Regarding response times, Gemini demonstrates lower latency, while GPT-3.5 exhibits the highest response time on the MotionSense dataset. However, for the WISDM dataset, the response times for GPT-3.5 and GPT-4 are quite similar. Overall, response time is influenced by the amount of text produced, which is closely related to the number of characters generated in each response. In the one-shot setting, both ChatGPT-4 and Gemini successfully detect and sanitize the poisoned labels across all datasets, effectively handling inter-class similarities and differences.

\subsubsection{Few-shot setting:}
In few-shot learning, we send three sensor examples corresponding to each activity, followed by sensor data from a poisoned activity to detect and sanitize the instance. Table \ref{tab:comparison_for_all} conclusively shows that the detection and sanitization of poisoning attacks in both the MotionSense and HHAR and WISDM datasets achieve a score of 1.0. This indicates that all LLMs (ChatGPT-3.5, ChatGPT-4, and Gemini) are capable of accurately identifying data poisoning attacks across activities, considering inter-class similarities and differences, related to the outcomes in one-shot learning. Moreover, few-shot learning exhibits lower communication costs and response times compared to zero-shot and one-shot learning for the MotionSense and HHAR datasets, as detailed in Fig. \ref{fig:response} and Fig. \ref{fig:communicationcost}. However, for the WISDM dataset, it shows higher communication costs for ChatGPT-3.5 and ChatGPT-4, as illustrated in Fig. \ref{fig:commcostW}. Additionally, the response time is higher for Gemini on the WISDM dataset, as shown in Fig. \ref{fig:responseTimeW}.

\begin{table}
  \caption{Comparison of Detection and Label Sanitization Among ChatGPT 3.5/4 and Gemini for Inter-Class Similarities in MotionSense Dataset in Zero-shot learning}
  \label{tab:comparison-inter-class-similarities-zero-shot-motionsense}
  \centering
  \resizebox{\columnwidth}{!}{%
    \begin{tabular}{
      >{\columncolor{green!20}}p{2.2cm}
      >{\columncolor{red!20}}p{2.6cm}|
      p{1.7cm} p{2.7cm}|
      p{1.7cm} p{2.7cm}|
      p{1.7cm} p{2.7cm}
    }
      \toprule
      \textbf{Actual Label} & \textbf{Poisoned Label} 
      & \multicolumn{2}{c|}{\textbf{GPT 3.5}} 
      & \multicolumn{2}{c|}{\textbf{GPT 4}} 
      & \multicolumn{2}{c}{\textbf{Gemini}} \\
      \cline{3-8}
      & & \textbf{Detection} & \textbf{Label Sanitization}
        & \textbf{Detection} & \textbf{Label Sanitization}
        & \textbf{Detection} & \textbf{Label Sanitization} \\
      \midrule

Standing   &  Sitting         &  Yes &  Standing          &  Yes &  Standing           &  Yes &  Standing            \\ 
Sitting    &  Standing        &  Yes & \fcolorbox{red!30}{red!30}{Sitting/Lying Down} &  Yes &  Sitting        &  Yes & \fcolorbox{red!30}{red!30}{Downstairs/Upstairs}  \\ 
Upstairs   &  Downstairs      & \fcolorbox{red!30}{red!30}{No}  & \fcolorbox{red!30}{red!30}{N/A}         &  Yes & \fcolorbox{red!30}{red!30}{Walking}            &  Yes & \fcolorbox{red!30}{red!30}{Jogging/Walking}  \\ 
Downstairs &  Upstairs        &  Yes & \fcolorbox{red!30}{red!30}{Walking}           &  Yes &  Downstairs         &  Yes &  Downstairs           \\ 
Upstairs   &  Jogging         &  Yes & \fcolorbox{red!30}{red!30}{Walking}   &  Yes &  Upstairs    & \fcolorbox{red!30}{red!30}{No} & \fcolorbox{red!30}{red!30}{N/A}        \\ 
Downstairs &  Jogging         &  Yes & \fcolorbox{red!30}{red!30}{Walking}   &  Yes &  Downstairs    &  Yes & \fcolorbox{red!30}{red!30}{Downstairs/Upstairs}       \\ 
Jogging    &  Upstairs        &  Yes & \fcolorbox{red!30}{red!30}{Downstairs}   &  Yes &  Jogging    &  Yes & \fcolorbox{red!30}{red!30}{Walking/Standing}        \\ 
Jogging    &  Downstairs      &  Yes & \fcolorbox{red!30}{red!30}{Walking/Jogging}   &  Yes &  Jogging    &  Yes & \fcolorbox{red!30}{red!30}{Walking/Standing}        \\ 
Jogging    &  Walking         &  Yes & \fcolorbox{red!30}{red!30}{Downstairs}         &  Yes &  Jogging            &  Yes & \fcolorbox{red!30}{red!30}{Standing}            \\ 
Walking    &  Jogging         &  Yes &  Walking            &  Yes &  Walking            &  Yes &  \fcolorbox{red!30}{red!30}{Walking/Standing}  \\ 
Walking    &  Upstairs        & \fcolorbox{red!30}{red!30}{No}  & \fcolorbox{red!30}{red!30}{N/A}           &  Yes &  Walking            &  Yes & \fcolorbox{red!30}{red!30}{Walking/Standing}  \\ 
Upstairs   &  Walking         &  Yes & \fcolorbox{red!30}{red!30}{Jogging}            &  Yes &  Upstairs           &  Yes & \fcolorbox{red!30}{red!30}{Jogging}            \\
Walking    &  Downstairs      & \fcolorbox{red!30}{red!30}{No}  & \fcolorbox{red!30}{red!30}{N/A}         &  Yes &  Walking            &  Yes & \fcolorbox{red!30}{red!30}{Walking/Standing}  \\ 
Downstairs &  Walking         &  Yes & \fcolorbox{red!30}{red!30}{Jogging}            &  Yes &  Downstairs         &  Yes & \fcolorbox{red!30}{red!30}{Downstairs/Upstairs} \\ 
      \hline
    \end{tabular}
  }
\end{table}

\begin{table}[t]
  \caption{Comparison of Detection and Label Sanitization in ChatGPT-3.5/4 and Gemini for Inter-Class Difference in the MotionSense Dataset in Zero-shot learning}
  \label{tab:comparison-motionsense-difference-zero}
  \centering
  \resizebox{\columnwidth}{!}{%
    \begin{tabular}{>{\columncolor{green!20}}p{2.5cm} >{\columncolor{red!20}}p{2.9cm}| p{2.2cm}p{3.7cm}| p{2.2cm} p{3.7cm}| p{2.2cm} p{3.7cm}}
      \toprule
      \textbf{Actual Label} & \textbf{Poisoned Label} & \multicolumn{2}{c|}{\textbf{ChatGPT-3.5}} & \multicolumn{2}{c|}{\textbf{ChatGPT-4}} & \multicolumn{2}{c}{\textbf{Gemini}} \\ 
      \cline{3-8}
                 &                 & \textbf{Detection} & \textbf{Label Sanitization} & \textbf{Detection} & \textbf{Label Sanitization} & \textbf{Detection} & \textbf{Label Sanitization} \\
      \midrule
       Standing   &  Walking         &  Yes &  Standing          &  Yes &  Standing           &  Yes &  Standing            \\ 
       Standing   &  Jogging         &  Yes &  Standing          &  Yes &  Standing           &  Yes &  Standing            \\ 
       Standing   &  Upstairs        &  Yes &  Standing          &  Yes &  Standing           &  Yes &  Standing            \\ 
       Standing   &  Downstairs      &  Yes &  Standing          &  Yes &  Standing           &  Yes &  Standing            \\ 
       Walking    &  Standing        &  Yes & \fcolorbox{red!30}{red!30}{Downstairs}  &  Yes &  Walking            &  \fcolorbox{red!30}{red!30}{No}  & \fcolorbox{red!30}{red!30}{N/A}            \\ 
       Jogging    &  Standing        &  Yes &  \fcolorbox{red!30}{red!30}{Walking/Jogging} &  Yes &  Jogging            & \fcolorbox{red!30}{red!30}{No} & \fcolorbox{red!30}{red!30}{N/A} \\ 
       Upstairs   &  Standing        &  Yes &  \fcolorbox{red!30}{red!30}{Jogging}           &  Yes &  Upstairs           &  Yes & \fcolorbox{red!30}{red!30}{Jogging/Walking}  \\ 
       Downstairs &  Standing        &  Yes & \fcolorbox{red!30}{red!30}{ Walking}           &  Yes &  Downstairs         &  Yes & \fcolorbox{red!30}{red!30}{Downstairs/Upstairs} \\ 
       Sitting    &  Walking         &  Yes & \fcolorbox{red!30}{red!30}{Sitting/Standing} &  Yes &  Sitting          &  Yes & \fcolorbox{red!30}{red!30}{Downstairs/Upstairs} \\ 
       Sitting    &  Jogging         &  Yes & \fcolorbox{red!30}{red!30}{Standing}          &  Yes &  Sitting            &  Yes &  Sitting            \\ 
       Sitting    &  Upstairs        &  Yes & \fcolorbox{red!30}{red!30}{Sitting/Standing} &  Yes &  Sitting          &  Yes & \fcolorbox{red!30}{red!30}{Walking or Jogging} \\ 
       Sitting    &  Downstairs      &  Yes & \fcolorbox{red!30}{red!30}{Sitting/Standing} &  Yes &  Sitting          &  Yes & \fcolorbox{red!30}{red!30}{Jogging/Walking}  \\ 
       Walking    &  Sitting         &  Yes & \fcolorbox{red!30}{red!30}{ Upstairs}    &  Yes &  Walking            &  Yes & \fcolorbox{red!30}{red!30}{Walking/Standing} \\ 
       Jogging    &  Sitting         &  Yes & \fcolorbox{red!30}{red!30}{Walking/Jogging} &  Yes &  Jogging            &  Yes & \fcolorbox{red!30}{red!30}{Walking/Standing} \\ 
       Upstairs   &  Sitting         &  Yes & \fcolorbox{red!30}{red!30}{ Jogging}           &  Yes &  Upstairs           &  Yes  & \fcolorbox{red!30}{red!30}{Jogging}            \\ 
       Downstairs &  Sitting         &  Yes & \fcolorbox{red!30}{red!30}{Walking}           &  Yes &  Downstairs         &  Yes & \fcolorbox{red!30}{red!30}{Downstairs/Upstairs} \\ \hline
    \end{tabular}%
  }
\end{table}

\begin{table}
  \caption{Comparison of Detection and Label Sanitization in ChatGPT-3.5/4 and Gemini for Inter-Class Similarities in HHAR Dataset in Zero-shot learning}
  \vspace{-5pt}
  \label{tab:comparison-inter-class-similarities-zero-shot-hhar}
  \centering

  \resizebox{\columnwidth}{!}{%
    \begin{tabular}{>{\columncolor{green!20}}p{2.5cm} >{\columncolor{red!20}}p{2.9cm}| p{2.2cm}p{3.7cm}| p{2.2cm} p{3.7cm}| p{2.2cm} p{3.7cm}}
      \toprule
      \textbf{Actual Label} & \textbf{Poisoned Label} & \multicolumn{2}{c|}{\textbf{ChatGPT-3.5}} & \multicolumn{2}{c|}{\textbf{ChatGPT-4}} & \multicolumn{2}{c}{\textbf{Gemini}} \\ 
      \cline{3-8}
                 &                 & \textbf{Detection} & \textbf{Label Sanitization} & \textbf{Detection} & \textbf{Label Sanitization} & \textbf{Detection} & \textbf{Label Sanitization} \\
      \midrule
  Standing &  Sitting & \fcolorbox{red!30}{red!30}{No} & \fcolorbox{red!30}{red!30}{N/A} & \fcolorbox{red!30}{red!30}{No} & \fcolorbox{red!30}{red!30}{N/A} &  Yes & \fcolorbox{red!30}{red!30}{ Stairsdown} \\

 Sitting &  Standing & \fcolorbox{red!30}{red!30}{No} & \fcolorbox{red!30}{red!30}{N/A} &  Yes &  Sitting & \fcolorbox{red!30}{red!30}{No} & \fcolorbox{red!30}{red!30}{N/A} \\

 Stairsup &  Stairsdown &  Yes & \fcolorbox{red!30}{red!30}{Sitting} &  Yes &  Stairsup &  Yes & \fcolorbox{red!30}{red!30}{Sitting/Standing} \\

 Stairsdown &  Stairsup &  Yes & \fcolorbox{red!30}{red!30}{Walking} &  Yes &  Stairsdown &  Yes &  Stairsdown \\

 Stairsup &  Biking &  Yes & \fcolorbox{red!30}{red!30}{Standing} &  Yes & \fcolorbox{red!30}{red!30}{Standing} &  Yes & \fcolorbox{red!30}{red!30}{Sitting/Standing} \\

 Stairsdown &  Biking &  Yes & \fcolorbox{red!30}{red!30}{Walking} &  Yes &  Stairsdown &  Yes & \fcolorbox{red!30}{red!30}{Stairsdown/Upstairs} \\

 Biking &  Stairsup &  Yes & \fcolorbox{red!30}{red!30}{Standing} &  Yes &  Biking &  Yes & \fcolorbox{red!30}{red!30}{Stairsdown} \\

 Biking &  Stairsdown &  Yes & \fcolorbox{red!30}{red!30}{Standing} &   Yes &  Biking &  Yes & \fcolorbox{red!30}{red!30}{Standing} \\

 Biking &  Walking &   Yes & \fcolorbox{red!30}{red!30}{Stairsdown} &  Yes &  Biking &  Yes & \fcolorbox{red!30}{red!30}{Standing} \\

 Walking &  Biking & \fcolorbox{red!30}{red!30}{No} & \fcolorbox{red!30}{red!30}{N/A} &  Yes &  Walking &  Yes & \fcolorbox{red!30}{red!30}{Standing} \\

 Walking &  Stairsup &  Yes & \fcolorbox{red!30}{red!30}{Standing} &   Yes &  Walking & \fcolorbox{red!30}{red!30}{No} & \fcolorbox{red!30}{red!30}{N/A} \\

 Stairsup &  Walking & \fcolorbox{red!30}{red!30}{No} & \fcolorbox{red!30}{red!30}{N/A} &  Yes &  Stairsup &  Yes &  Stairsup \\

 Walking &  Stairsdown &  Yes & \fcolorbox{red!30}{red!30}{Sitting or Standing} &  Yes & \fcolorbox{red!30}{red!30}{Walking or Stairsup} &  Yes & \fcolorbox{red!30}{red!30}{Standing} \\
 Stairsdown &  Walking &  Yes &  Stairsdown &  Yes &  Stairsdown &  Yes &  Stairsdown \\ \hline
 
    \end{tabular}%
  }
\end{table}

\begin{table}[t]
  \caption{Comparison of Detection and Label Sanitization in ChatGPT-3.5/4 and Gemini for Inter-Class Difference in the HHAR Dataset in Zero-shot learning}
  \label{tab:comparison-interclass-difference-hhar-zero}
 \centering
  \resizebox{\columnwidth}{!}{%
    \begin{tabular}{>{\columncolor{green!20}}p{2.5cm} >{\columncolor{red!20}}p{2.9cm}| p{2.2cm}p{3.7cm}| p{2.2cm} p{3.7cm}| p{2.2cm} p{3.7cm}}
      \toprule
      \textbf{Actual Label} & \textbf{Poisoned Label} & \multicolumn{2}{c|}{\textbf{ChatGPT-3.5}} & \multicolumn{2}{c|}{\textbf{ChatGPT-4}} & \multicolumn{2}{c}{\textbf{Gemini}} \\ 
      \cline{3-8}
                 &                 & \textbf{Detection} & \textbf{Label Sanitization} & \textbf{Detection} & \textbf{Label Sanitization} & \textbf{Detection} & \textbf{Label Sanitization} \\
      \midrule
 Standing &  Walking &  Yes &  Standing &  Yes &  Standing &  Yes & \fcolorbox{red!30}{red!30}{Stairsdown} \\
       Standing &  Biking &  Yes &  Standing &  Yes &  Standing &  Yes &  Standing \\
       Standing &  Stairsup &  Yes &  Standing &  Yes &  Standing &  Yes & \fcolorbox{red!30}{red!30}{Stairsdown} \\
       Standing &  Stairsdown &  Yes & \fcolorbox{red!30}{red!30} {Sitting} &  Yes &  Standing &  Yes &  Standing \\
       Walking &  Standing &  Yes & \fcolorbox{red!30}{red!30}{Stairsup/Stairsdown} &  Yes &  Walking &  Yes & \fcolorbox{red!30}{red!30}{Stairsup} \\
       Biking &  Standing & \fcolorbox{red!30}{red!30}{No} & \fcolorbox{red!30}{red!30}{N/A} &  Yes &  Biking &  Yes & \fcolorbox{red!30}{red!30}{Stairsdown} \\

       Stairsup &  Standing &  Yes &  Stairsup &  Yes &  Stairsup &  Yes &  Stairsup \\
       Stairsdown &  Standing &  Yes & \fcolorbox{red!30}{red!30}{Stairsup} &  Yes &  Stairsdown &  Yes &  Stairsdown \\
       Sitting &  Walking &  Yes & \fcolorbox{red!30}{red!30}{Standing} &  Yes &  Sitting &  Yes & \fcolorbox{red!30}{red!30}{Standing} \\
       Sitting &  Biking &  Yes & \fcolorbox{red!30}{red!30}{Sitting/Standing} &  Yes &  Sitting &  Yes &  Sitting \\
       Sitting &  Stairsup &  Yes & \fcolorbox{red!30}{red!30}{Sitting/Standing} &  Yes &  Sitting &  Yes &  Standing \\
       Sitting &  Stairsdown &  Yes &  Sitting &  Yes &  Sitting &  Yes & \fcolorbox{red!30}{red!30}{Standing} \\
       Walking &  Sitting &  Yes &  Walking &  Yes &  Walking &  Yes & \fcolorbox{red!30}{red!30}{Standing} \\
       Biking &  Sitting &  Yes & \fcolorbox{red!30}{red!30}{Walking} &  Yes &  Biking &  Yes & \fcolorbox{red!30}{red!30}{Standing} \\
       Stairsup &  Sitting &  Yes & \fcolorbox{red!30}{red!30}{Standing} &  Yes &  Stairsup &  Yes & \fcolorbox{red!30}{red!30}{Sitting/Standing} \\
       Stairsdown &  Sitting &  Yes & \fcolorbox{red!30}{red!30}{Walking} &  Yes &  Stairsdown &  Yes & \fcolorbox{red!30}{red!30}{Standing} \\
      \hline
    \end{tabular}
  }
\end{table}

\begin{table}[h]
\centering
\caption{{Comparison of Detection and Label Sanitization in ChatGPT-3.5/4 and Gemini for Inter-Class Similarities in WISDM Dataset in Zero-shot learning}}
\label{tab:detection_sanitization_interclass_similaraties_WISDM}
\resizebox{\columnwidth}{!}{%
\begin{tabular}{>{\columncolor{green!20}}p{2.5cm}>{\columncolor{red!20}}p{3.0cm}|p{2.2cm}p{3.7cm}|p{2.2cm}p{3.7cm}|p{2.2cm}p{3.7cm}}
\toprule
\textbf{Actual Label} & \textbf{Poisoned Label} & \multicolumn{2}{c}{\textbf{ChatGPT-3.5}} & \multicolumn{2}{c}{\textbf{ChatGPT-4}} & \multicolumn{2}{c}{\textbf{Gemini}} \\
\cline{3-8}
& & \textbf{Detection} & \textbf{Label Sanitization} & \textbf{Detection} & \textbf{Label Sanitization} & \textbf{Detection} & \textbf{Label Sanitization} \\
\midrule
Standing   & Sitting     & Yes & \fcolorbox{red!30}{red!30}{Walking}                & Yes & Standing                 & Yes & \fcolorbox{red!30}{red!30}{Jogging} \\
Sitting    & Standing    & Yes & \fcolorbox{red!30}{red!30}{Walking}                & Yes & \fcolorbox{red!30}{red!30}{Sitting or Standing}      & Yes & Sitting \\
Walking    & Jogging     & Yes & \fcolorbox{red!30}{red!30}{Sitting}                & Yes & \fcolorbox{red!30}{red!30}{Walking or Jogging}       & Yes & Upstairs \\
Walking    & Downstairs  & Yes & \fcolorbox{red!30}{red!30}{Upstairs}               & Yes & Walking                  & Yes & \fcolorbox{red!30}{red!30}{Upstairs} \\
Walking    & Upstairs    & Yes & \fcolorbox{red!30}{red!30}{Downstairs}             & Yes & Walking                  & Yes & \fcolorbox{red!30}{red!30}{Downstairs} \\
Jogging    & Walking     & Yes & \fcolorbox{red!30}{red!30}{Standing}               & Yes & Jogging                  & Yes & \fcolorbox{red!30}{red!30}{Sitting or Standing} \\
Jogging    & Downstairs  & Yes & \fcolorbox{red!30}{red!30}{Upstairs}               & Yes & Jogging                  & \fcolorbox{red!30}{red!30}{No}  & \fcolorbox{red!30}{red!30}{N/A} \\
Jogging    & Upstairs    & Yes & \fcolorbox{red!30}{red!30}{Downstairs}             & Yes & Jogging                  & Yes & Jogging \\
Downstairs & Upstairs    & Yes & Downstairs             & Yes & Downstairs               & Yes & \fcolorbox{red!30}{red!30}{Jogging} \\
Downstairs & Jogging     & Yes & \fcolorbox{red!30}{red!30}{Standing}               & Yes & Downstairs               & Yes & \fcolorbox{red!30}{red!30}{Walking} \\
Downstairs & Walking     & Yes & \fcolorbox{red!30}{red!30}{Sitting}                & Yes & Downstairs               & Yes & Jogging \\
Upstairs   & Downstairs  & Yes & Upstairs               & Yes & \fcolorbox{red!30}{red!30}{Upstairs or Downstairs}   & Yes & Upstairs \\
Upstairs   & Jogging     & Yes & \fcolorbox{red!30}{red!30}{Standing}               & Yes & Upstairs                 & Yes & \fcolorbox{red!30}{red!30}{Walking} \\
Upstairs   & Walking     & Yes & \fcolorbox{red!30}{red!30}{Standing}               & Yes & Upstairs                 & Yes & \fcolorbox{red!30}{red!30}{Jogging} \\
\bottomrule
\end{tabular}%
}
\end{table}

\begin{table}[h]
\centering
\caption{{Comparison of Detection and Label Sanitization in ChatGPT-3.5, ChatGPT-4, and Gemini for Inter-Class Difference in WISDM Dataset in Zero-shot Learning}}
\label{tab:detection_sanitization_difference_WISDM}
\resizebox{\columnwidth}{!}{%
\begin{tabular}{>{\columncolor{green!20}}p{2.5cm}>{\columncolor{red!20}}p{3.0cm}|p{2.2cm}p{3.7cm}|p{2.2cm}p{3.7cm}|p{2.2cm}p{3.7cm}}
\toprule
\textbf{Actual Label} & \textbf{Poisoned Label} & \multicolumn{2}{c}{\textbf{ChatGPT-3.5}} & \multicolumn{2}{c}{\textbf{ChatGPT-4}} & \multicolumn{2}{c}{\textbf{Gemini}} \\
\cline{3-8}
& & \textbf{Detection} & \textbf{Label Sanitization} & \textbf{Detection} & \textbf{Label Sanitization} & \textbf{Detection} & \textbf{Label Sanitization} \\
\midrule
Standing   & Downstairs & Yes & \fcolorbox{red!30}{red!30}{Upstairs}     & Yes & Standing     & Yes & \fcolorbox{red!30}{red!30}{Walking} \\
Standing   & Upstairs   & Yes & \fcolorbox{red!30}{red!30}{Downstairs}   & Yes & Standing     & Yes & Jogging \\
Standing   & Walking   & \fcolorbox{red!30}{red!30}{No}  & \fcolorbox{red!30}{red!30}{N/A}          & Yes & Standing     & Yes & \fcolorbox{red!30}{red!30}{Upstairs} \\
Standing   & Jogging   & Yes & \fcolorbox{red!30}{red!30}{Walking}      & Yes & Standing     & Yes & \fcolorbox{red!30}{red!30}{Walking} \\
Sitting    & Downstairs & Yes & \fcolorbox{red!30}{red!30}{Upstairs}     & Yes & Sitting      & Yes & \fcolorbox{red!30}{red!30}{Standing} \\
Sitting    & Upstairs   & Yes & \fcolorbox{red!30}{red!30}{Downstairs}   & Yes & Sitting      & Yes & Sitting \\
Sitting    & Walking    & Yes & Sitting      & Yes & Sitting      & Yes & Sitting \\
Sitting    & Jogging    & Yes & Sitting      & Yes & Sitting      & Yes & \fcolorbox{red!30}{red!30}{Standing} \\
Walking    & Standing   & Yes & Walking      & Yes & \fcolorbox{red!30}{red!30}{Jogging}      & Yes & \fcolorbox{red!30}{red!30}{Jogging} \\
Walking    & Sitting    & Yes & \fcolorbox{red!30}{red!30}{Jogging}      & Yes & Walking      & Yes & \fcolorbox{red!30}{red!30}{Jogging} \\
Jogging    & Standing   & Yes & \fcolorbox{red!30}{red!30}{Sitting}      & Yes & Jogging      & Yes & Jogging \\
Jogging    & Sitting    & Yes & \fcolorbox{red!30}{red!30}{Standing}     & Yes & Jogging      & Yes & \fcolorbox{red!30}{red!30}{Sitting or Standing} \\
Downstairs & Standing   & Yes & \fcolorbox{red!30}{red!30}{Sitting}      & Yes & Downstairs   & Yes & \fcolorbox{red!30}{red!30}{Upstairs} \\
Downstairs & Sitting    & Yes & \fcolorbox{red!30}{red!30}{Jogging}      & Yes & Downstairs   & Yes & \fcolorbox{red!30}{red!30}{Jogging} \\
Upstairs   & Standing   & Yes & \fcolorbox{red!30}{red!30}{Downstairs}   & Yes & Upstairs     & Yes & \fcolorbox{red!30}{red!30}{Jogging} \\
Upstairs   & Sitting    & Yes & \fcolorbox{red!30}{red!30}{Jogging}      & Yes & Upstairs     & Yes & \fcolorbox{red!30}{red!30}{Jogging} \\
\bottomrule
\end{tabular}%
}
\end{table}

\begin{table}[h]
  \caption{Comparison of Detection and Label Sanitization in ChatGPT-3.5/4 and Gemini for Inter-Class Similarities in motionsense Dataset in One-shot learning}
  \label{tab:comparison_similar}
  \centering
  \resizebox{\columnwidth}{!}{%
    \begin{tabular}{>{\columncolor{green!20}}p{2.5cm} >{\columncolor{red!20}}p{2.9cm}| p{2.2cm}p{3.7cm}| p{2.2cm} p{3.7cm}| p{2.2cm} p{3.7cm}}
      \toprule
      \textbf{Actual Label} & \textbf{Poisoned Label} & \multicolumn{2}{c|}{\textbf{ChatGPT 3.5}} & \multicolumn{2}{c|}{\textbf{ChatGPT 4}} & \multicolumn{2}{c}{\textbf{Gemini}} \\ 
      \cline{3-8}
                 &                 & \textbf{Detection} & \textbf{Label Sanitization} & \textbf{Detection} & \textbf{Label Sanitization} & \textbf{Detection} & \textbf{Label Sanitization} \\ 
      \midrule
       Standing &  Sitting & \fcolorbox{red!30}{red!30}{ No} & \fcolorbox{red!30}{red!30}{ N/A} &  Yes &  Standing &  Yes &  Standing \\
       Sitting &  Standing &  Yes &  Sitting &  Yes &  Sitting &  Yes &  Sitting \\
       Upstairs &  Downstairs &  Yes &  Upstairs &  Yes &  Walking &  Yes &  Walking \\
       Downstairs &  Upstairs &  Yes &  Downstairs &  Yes &  Downstairs &  Yes &  Downstairs \\
       Upstairs &  Jogging &  Yes &  Upstairs &  Yes &  Upstairs &  Yes &  Upstairs \\
       Downstairs &  Jogging &  Yes &  Downstairs &  Yes &  Downstairs &  Yes &  Downstairs \\
       Jogging &  Upstairs &  Yes &  Jogging &  Yes &  Jogging &  Yes &  Jogging \\
       Jogging &  Downstairs &  Yes &  Jogging &  Yes &  Jogging &  Yes &  Jogging \\
       Jogging &  Walking &  Yes &  Downstairs &  Yes &  Jogging &  Yes &  Jogging \\
       Walking &  Jogging &  Yes &  Walking &  Yes &  Walking &  Yes &  Walking \\
       Walking &  Upstairs &  Yes &  Walking &  Yes &  Walking &  Yes &  Walking \\
       Upstairs &  Walking &  Yes &  Upstairs &  Yes &  Upstairs &  Yes &  Upstairs \\
       Walking &  Downstairs &  Yes &  Walking &  Yes &  Walking &  Yes &  Walking \\
       Downstairs &  Walking &  Yes &  Downstairs &  Yes &  Downstairs &  Yes &  Downstairs \\
      \hline
    \end{tabular}%
  }
\end{table}

\begin{table}
  \caption{Comparison of Detection and Label Sanitization in ChatGPT-3.5/4 and Gemini for inter-class difference in motionsense dataset One-shot learning}
  \label{tab:comparison_difference}
  \centering
  \resizebox{\columnwidth}{!}{%
    \begin{tabular}{>{\columncolor{green!20}}p{2.5cm} >{\columncolor{red!20}}p{2.9cm}| p{2.2cm}p{3.7cm}| p{2.2cm} p{3.7cm}| p{2.2cm} p{3.7cm}}
      \toprule
      \textbf{Actual Label} & \textbf{Poisoned Label} & \multicolumn{2}{c}{\textbf{ChatGPT-3.5}} & \multicolumn{2}{c}{\textbf{ChatGPT-4}} & \multicolumn{2}{c}{\textbf{Gemini}} \\ 
      \cline{3-8}
      & & \textbf{Detection} & \textbf{Label Sanitization} & \textbf{Detection} & \textbf{Label Sanitization} & \textbf{Detection} & \textbf{Label Sanitization} \\ 
      \midrule
       Standing &  Walking &  Yes &  Standing &  Yes &  Standing &  Yes &  Standing \\
       Standing &  Jogging &  Yes &  Standing &  Yes &  Standing &  Yes &  Standing \\
       Standing &  Upstairs &  Yes &  Standing &  Yes &  Standing &  Yes &  Standing \\
       Standing &  Downstairs &  Yes &  Standing &  Yes &  Standing &  Yes &  Standing \\
       Walking &  Standing &  Yes &  Walking &  Yes &  Walking &  Yes &  Walking\\
       Walking &  Sitting &  Yes &  Walking &  Yes &  Walking &  Yes &  Walking \\
       Jogging &  Standing &  Yes &  Jogging &  Yes &  Jogging &  Yes &  Jogging \\
       Jogging &  Sitting &  Yes &  Jogging &  Yes &  Jogging &  Yes &  Jogging \\
       Upstairs &  Standing &  Yes &  Upstairs &  Yes &  Upstairs &  Yes &  Upstairs \\
       Upstairs &  Sitting &  Yes &  Upstairs &  Yes &  Upstairs &  Yes &  Upstairs \\
       Downstairs &  Sitting &  Yes &  Downstairs &  Yes &  Downstairs &  Yes &  Downstairs \\
       Downstairs &  Standing &  Yes &  Downstairs &  Yes &  Downstairs &  Yes &  Downstairs \\
       Sitting &  Walking &  Yes &  Sitting &  Yes &  Sitting &  Yes &  Sitting \\
       Sitting &  Jogging &  Yes &  Sitting &  Yes &  Sitting &  Yes &  Sitting \\
       Sitting &  Upstairs &  Yes &  Sitting &  Yes &  Sitting &  Yes &  Sitting \\
       Sitting &  Downstairs &  Yes &  Sitting &  Yes &  Sitting &  Yes &  Sitting \\
      \bottomrule
    \end{tabular}%
  }
\end{table}

\begin{table}[ht]
\centering
\caption{Comparison of Detection and Label Sanitization in ChatGPT-3.5/4 and Gemini for inter-class similarities in the HHAR dataset in one-shot learning.}
\label{tab:comparison_difference}
\resizebox{\columnwidth}{!}{%
\begin{tabular}{>{\LARGE\columncolor{green!20}}p{2.5cm} >{\LARGE\columncolor{red!20}}p{2.9cm}| >{\LARGE}p{2.2cm} >{\LARGE}p{3.7cm}| >{\LARGE}p{2.2cm} >{\LARGE}p{3.7cm}| >{\LARGE}p{2.2cm} >{\LARGE}p{3.7cm}}
\toprule
\textbf{Actual Label} & \textbf{Poisoned Label} & \multicolumn{2}{c}{\LARGE\textbf{ChatGPT-3.5}} & \multicolumn{2}{c}{\LARGE\textbf{ChatGPT-4}} & \multicolumn{2}{c}{\LARGE\textbf{Gemini}} \\ 
\cline{3-8}
 & & \textbf{Detection} & \textbf{Label Sanitization} & \textbf{Detection} & \textbf{Label Sanitization} & \textbf{Detection} & \textbf{Label Sanitization} \\ 
\midrule
Standing  & Sitting   & Yes & Standing & Yes & Standing & Yes & Standing \\
Sitting   & Standing  & Yes & Sitting  & Yes & Sitting  & Yes & Sitting \\
Stairsup  & Stairsdown& Yes & Stairsup & Yes & Stairsup & Yes & Stairsup \\
Stairsup  & Biking    & Yes & Stairsup & Yes & Stairsup & Yes & Stairsup \\
Stairsup  & Walking   & Yes & Stairsup & Yes & Stairsup & Yes & Stairsup \\
Stairsdown& Stairsup  & Yes & Stairsdown & Yes & Stairsdown & Yes & Stairsdown \\
Stairsdown& Walking   & Yes & Stairsdown & Yes & Stairsdown & Yes & Stairsdown \\
Stairsdown& Biking    & Yes & Stairsdown & Yes & Stairsdown & Yes & Stairsdown \\
Biking    & Stairsup  & Yes & Biking    & Yes & Biking    & Yes & Biking \\
Biking    & Stairsdown& Yes & Biking    & Yes & Biking    & Yes & Biking \\
Biking    & Walking   & Yes & Biking    & Yes & Biking    & Yes & Biking \\
Walking   & Biking    & Yes & Walking   & Yes & Walking   & Yes & Walking \\
Walking   & Stairsup  & Yes & Walking   & Yes & Walking   & Yes & Walking \\
Walking   & Stairsdown& Yes & Walking   & Yes & Walking   & Yes & Walking \\
\bottomrule
\end{tabular}%
}
\end{table}

\begin{table}[h]
\centering
\caption{Comparison of Detection and Label Sanitization in ChatGPT-3.5/4 and Gemini for inter-class difference in the HHAR dataset in one-shot learning.}
\label{tab:comparison_Difference_HHAR}
\resizebox{\columnwidth}{!}{%
\begin{tabular}{>{\LARGE\columncolor{green!20}}p{2.5cm}>{\LARGE\columncolor{red!20}}p{3.0cm}|>{\LARGE}p{2.2cm}>{\LARGE}p{3.7cm}|>{\LARGE}p{2.2cm}>{\LARGE}p{3.7cm}|>{\LARGE}p{2.2cm}>{\LARGE}p{3.7cm}}
\toprule
\textbf{Actual Label} & \textbf{Poisoned Label} & \multicolumn{2}{c}{\LARGE\textbf{ChatGPT-3.5}} & \multicolumn{2}{c}{\LARGE\textbf{ChatGPT-4}} & \multicolumn{2}{c}{\LARGE\textbf{Gemini}} \\ 
\cline{3-8}
 & & \textbf{Detection} & \textbf{Label Sanitization} & \textbf{Detection} & \textbf{Label Sanitization} & \textbf{Detection} & \textbf{Label Sanitization} \\ 
\midrule
Standing & Walking & Yes & Standing & Yes & Standing & Yes & Standing \\
Standing & Biking & Yes & Standing & Yes & Standing & Yes & Standing \\
Standing & Stairsup & Yes & Standing & Yes & Standing & Yes & Standing \\
Standing & Stairsdown & Yes & Standing & Yes & Standing & Yes & Standing \\
Walking & Standing & Yes & Walking & Yes & Walking & Yes & Walking \\
Walking & Sitting & Yes & Walking & Yes & Walking & Yes & Walking \\
Biking & Standing & Yes & Biking & Yes & Biking & Yes & Biking \\
Biking & Sitting & Yes & Biking & Yes & Biking & Yes & Biking \\
Stairsup & Standing & Yes & \fcolorbox{red!20}{red!20}{Stairsup/ stairsdown} & Yes & Stairsup & Yes & Stairsup \\
Stairsup & Sitting & Yes & Stairsup & Yes & Stairsup & Yes & Stairsup \\
Stairsdown & Standing & Yes & Stairsdown & Yes & Stairsdown & Yes & Stairsdown \\
Stairsdown & Sitting & Yes & Stairsdown & Yes & Stairsdown & Yes & Stairsdown \\
Sitting & Walking & Yes & Sitting & Yes & Sitting & Yes & Sitting \\
Sitting & Biking & Yes & Sitting & Yes & Sitting & Yes & Sitting \\
Sitting & Stairsup & Yes & \fcolorbox{red!20}{red!20}{Sitting/Standing} & Yes & Sitting & Yes & Sitting \\
Sitting & Stairsdown & Yes & Sitting & Yes & Sitting & Yes & Sitting \\
\hline
\end{tabular}%
}
\end{table}

\begin{table}[h]
\centering
\caption{{Comparison of Detection and Label Sanitization in ChatGPT-3.5/4 and Gemini for inter-class similarities in the WISDM dataset using one-shot learning.}}
\label{tab:comparison_Distinct_HHAR}
\resizebox{\columnwidth}{!}{%
\begin{tabular}{>{\LARGE\columncolor{green!20}}p{2.5cm}>{\LARGE\columncolor{red!20}}p{3.0cm}|>{\LARGE}p{2.2cm}>{\LARGE}p{3.7cm}|>{\LARGE}p{2.2cm}>{\LARGE}p{3.7cm}|>{\LARGE}p{2.2cm}>{\LARGE}p{3.7cm}}
\toprule
\textbf{Actual Label} & \textbf{Poisoned Label} & \multicolumn{2}{c}{\LARGE\textbf{ChatGPT-3.5}} & \multicolumn{2}{c}{\LARGE\textbf{ChatGPT-4}} & \multicolumn{2}{c}{\LARGE\textbf{Gemini}} \\
\cline{3-8}
 & & \textbf{Detection} & \textbf{Label Sanitization} & \textbf{Detection} & \textbf{Label Sanitization} & \textbf{Detection} & \textbf{Label Sanitization} \\
\midrule
Standing   & Sitting     & Yes & Standing   & Yes & Standing   & Yes & Standing \\
Sitting    & Standing    & Yes & Sitting    & Yes & Sitting    & Yes & Sitting \\
Walking    & Jogging     & Yes & Walking    & Yes & Walking    & Yes & Walking \\
Walking    & Downstairs  & Yes & Walking    & Yes & Walking    & Yes & Walking \\
Walking    & Upstairs    & Yes & Walking    & Yes & Walking    & Yes & Walking \\
Jogging    & Walking     & Yes & Jogging    & Yes & Jogging    & Yes & Jogging \\
Jogging    & Downstairs  & Yes & Jogging    & Yes & Jogging    & Yes & Jogging \\
Jogging    & Upstairs    & Yes & Jogging    & Yes & Jogging    & Yes & Jogging \\
Downstairs & Upstairs    & Yes & Downstairs & Yes & Downstairs & Yes & Downstairs \\
Downstairs & Jogging     & Yes & Downstairs & Yes & Downstairs & Yes & Downstairs \\
Downstairs & Walking     & Yes & Downstairs & Yes & Downstairs & Yes & Downstairs \\
Upstairs   & Downstairs  & Yes & Upstairs   & Yes & Upstairs   & Yes & Upstairs \\
Upstairs   & Jogging     & Yes & Upstairs   & Yes & Upstairs   & Yes & Upstairs \\
Upstairs   & Walking     & Yes & Upstairs   & Yes & Upstairs   & Yes & Upstairs \\
\bottomrule
\end{tabular}%
}
\end{table}

\begin{table}[h]
\centering
\caption{{Comparison of Detection and Label Sanitization in ChatGPT-3.5/4 and Gemini for inter-class difference in the WISDM dataset using One-shot learning.}}
\label{tab:comparison_Consistent_HHAR}
\resizebox{\columnwidth}{!}{%
\begin{tabular}{>{\columncolor{green!20}}p{2.5cm}>{\columncolor{red!20}}p{3.0cm}|>{}p{2.2cm}>{}p{3.7cm}|>{}p{2.2cm}>{}p{3.7cm}|>{}p{2.2cm}>{}p{3.7cm}}
\toprule
\textbf{Actual Label} & \textbf{Poisoned Label} & \multicolumn{2}{c}{\textbf{ChatGPT-3.5}} & \multicolumn{2}{c}{\textbf{ChatGPT-4}} & \multicolumn{2}{c}{\textbf{Gemini}} \\
\cline{3-8}
 & & \textbf{Detection} & \textbf{Label Sanitization} & \textbf{Detection} & \textbf{Label Sanitization} & \textbf{Detection} & \textbf{Label Sanitization} \\
\midrule
Standing   & Downstairs & Yes & Standing   & Yes & Standing   & Yes & Standing \\
Standing   & Upstairs   & Yes & Standing   & Yes & Standing   & Yes & Standing \\
Standing   & Walking    & Yes & Standing   & Yes & Standing   & Yes & Standing \\
Standing   & Jogging    & Yes & Standing   & Yes & Standing   & Yes & Standing \\
Sitting    & Downstairs & Yes & Sitting    & Yes & Sitting    & Yes & Sitting \\
Sitting    & Upstairs   & Yes & Sitting    & Yes & Sitting    & Yes & Sitting \\
Sitting    & Walking    & Yes & Sitting    & Yes & Sitting    & Yes & Sitting \\
Sitting    & Jogging    & Yes & Sitting    & Yes & Sitting    & Yes & Sitting \\
Walking    & Standing   & Yes & Walking    & Yes & Walking    & Yes & Walking \\
Walking    & Sitting    & Yes & Walking    & Yes & Walking    & Yes & Walking \\
Jogging    & Standing   & Yes & Jogging    & Yes & Jogging    & Yes & Jogging \\
Jogging    & Sitting    & Yes & Jogging    & Yes & Jogging    & Yes & Jogging \\
Downstairs & Standing   & Yes & Downstairs & Yes & Downstairs & Yes & Downstairs \\
Downstairs & Sitting    & Yes & Downstairs & Yes & Downstairs & Yes & Downstairs \\
Upstairs   & Standing   & Yes & Upstairs   & Yes & Upstairs   & Yes & Upstairs \\
Upstairs   & Sitting    & Yes & Upstairs   & Yes & Upstairs   & Yes & Upstairs \\
\bottomrule
\end{tabular}%
}
\end{table}

\begin{table}[ht]
\centering
\caption{{Comparison of Poisoning Detection Accuracy (Acc.) and Sanitization Quality (SQ) of LLMs for MotionSense, HHAR, and WISDM Datasets in Zero-shot, One-shot, and Few-shot Learning.}}
\label{tab:comparison_accuracy_sq}
\scriptsize
\begin{tabular}{l|cccccc|cccccc|cccccc}
\toprule
\textbf{Model} 
& \multicolumn{6}{c|}{\textbf{MotionSense}} 
& \multicolumn{6}{c|}{\textbf{HHAR}} 
& \multicolumn{6}{c}{WISDM} \\
\cmidrule{2-19}
& \multicolumn{2}{c}{Zero-shot} & \multicolumn{2}{c}{One-shot} & \multicolumn{2}{c|}{Few-shot} 
& \multicolumn{2}{c}{Zero-shot} & \multicolumn{2}{c}{One-shot} & \multicolumn{2}{c|}{Few-shot} 
& \multicolumn{2}{c}{Zero-shot} & \multicolumn{2}{c}{One-shot} & \multicolumn{2}{c}{Few-shot} \\
\cmidrule{2-19}
& Acc. & SQ & Acc. & SQ & Acc. & SQ 
& Acc. & SQ & Acc. & SQ & Acc. & SQ 
& Acc. & SQ & Acc. & SQ & Acc. & SQ \\
\midrule
GPT 3.5   & 0.90 & 0.20 & 0.97 & 1.00    & 1.00    & 1.00    & 0.83 & 0.28 & 1.00    & 0.93 & 1.00    & 1.00    & 0.97 & 0.13 & 1.00 & 1.00    & 1.00    & 1.00    \\
GPT 4     & 1.00 & 0.97 & 1.00 & 1.00    & 1.00    & 1.00    & 0.97 & 0.90 & 1.00    & 1.00   & 1.00    & 1.00    & 1.00 & 0.83 & 1.00 & 1.00    & 1.00    & 1.00    \\
Gemini    & 0.90 & 0.26 & 1.00 & 1.00    & 1.00   & 1.00    & 0.93 & 0.32 & 1.00    & 1.00    & 1.00    & 1.00    & 0.97 & 0.16 & 1.00 & 1.00    & 1.00    & 1.00    \\
\bottomrule
\end{tabular}
\end{table}

\begin{table*}[ht]
\centering
\caption{{Comparison of Poisoning Detection Accuracy (Acc.) and Recall (Rec.) for KNN, LSTM, and TinyBERT Models across MotionSense, HHAR, and WISDM datasets in One-shot Learning}}
\label{tab:transposed_accuracy_recall}
\scriptsize
\setlength{\tabcolsep}{3pt}
\begin{tabular}{llcccccc|cccccc|cccccc}
\toprule
\textbf{Model} & \textbf{Metric} & \multicolumn{6}{c|}{\textbf{MotionSense}} & \multicolumn{6}{c|}{\textbf{HHAR}} & \multicolumn{6}{c}{\textbf{WISDM}} \\
\cmidrule(lr){3-8} \cmidrule(lr){9-14} \cmidrule(lr){15-20}
& & Stand & Sit & Up & Down & Jog & Walk & Stand & Sit & Up & Down & Bike & Walk & Stand & Sit & Up & Down & Jog & Walk \\
\midrule
KNN      & Acc. & 1.00 & 1.00 & 1.00 & 1.00 & 1.00 & 1.00 & 1.00 & 1.00 & 1.00 & 1.00 & 1.00 & 1.00 & 1.00 & 1.00 & 1.00 & 1.00 & 1.00 & 0.99 \\
KNN      & Rec. & 1.00 & 1.00 & 1.00 & 1.00 & 1.00 & 1.00 & 1.00 & 1.00 & 1.00 & 1.00 & 1.00 & 1.00 & 0.50 & 1.00 & 0.33 & 1.00 & 0.47 & 0.50 \\ \hline
LSTM     & Acc. & 1.00 & 1.00 & 0.93 & 0.98 & 0.93 & 0.90 & 1.00 & 1.00 & 0.73 & 0.51 & 1.00 & 0.27 & 1.00 & 1.00 & 0.90 & 0.60 & 1.00 & 1.00 \\
LSTM     & Rec. & 1.00 & 1.00 & 0.00 & 0.98 & 0.98 & 0.98 & 1.00 & 1.00 & 0.24 & 0.13 & 1.00 & 0.09 & 1.00 & 1.00 & 0.45 & 0.30 & 1.00 & 1.00 \\ \hline
{TinyBERT} & Acc. & 0.99 & 0.69 & 0.15 & 0.97 & 0.62 & 0.24 & 1.00 & 1.00 & 0.76 & 0.53 & 0.95 & 0.32 & 0.89 & 0.99 & 0.15 & 0.70 & 0.70 & 0.58 \\
{TinyBERT} & Rec. & 0.50 & 1.00 & 0.04 & 0.24 & 0.15 & 0.06 & 1.00 & 1.00 & 0.16 & 0.13 & 1.00 & 0.06 & 0.89 & 0.99 & 0.15 & 0.70 & 0.70 & 0.58\\
\bottomrule
\end{tabular}
\end{table*}

\subsection{Experiment with Traditional Methods}
We compare our proposed approach with KNN and LSTM, and TinyBert models on the same datasets (MotionSense and HHAR, and WISDM). This analysis includes 40,915 samples per activity, totaling 245,490 samples from all datasets. For the KNN model, we employ Scikit-learn’s GridSearchCV, testing odd values from 1 to 29 for the number of nearest neighbors to select the optimal parameter, with a 5-fold cross-validation (cv=5) for tuning. In the LSTM model, we implement one LSTM layer and one hidden layer using PyTorch, training the model over 100 epochs. The TinyBert model has 4 transformer layers and 312-dimensional hidden layers, and 12 attention heads; the total model is 14.5 million. In this evaluation, we compute accuracy and Sanitization Quality for each activity, enabling a comparative analysis of sanitization accuracy across KNN, LSTM, TinyBert, and LLMs. Accuracy and Sanitization Quality are calculated based on both true and poisoned labels.

The KNN models take approximately 11,010 seconds to train on the MotionSense dataset and 1,866 seconds on the HHAR dataset. The testing dataset is the same as LLMs 100 data, and the time is around 0.01 or 0.02 seconds for all models. The LSTM model requires about 1,713 seconds to train on the MotionSense dataset and 1,990 seconds on the HHAR dataset. These times are calculated approximately. The TinyBERT model takes around 438 seconds for training and 0.04 seconds for testing on the HHAR dataset, approximately 500 seconds for training and 0.02 seconds for testing on the MotionSense dataset, and 300 seconds for training and 0.02 seconds for testing on the WISDM dataset.

\begin{figure*}[ht]
    \centering
        \small
\hspace{0.5cm}
{\textcolor{blue}{\textbf{Standing}}}
\hspace{1.5cm}
{\textcolor{red}{\textbf{Sitting}}}
\hspace{1.3cm}
{\textcolor{cyan}{\textbf{Jogging}}}
\hspace{1.2cm}
{\textcolor{orange}{\textbf{Walking}}}
\hspace{1cm}
{\textcolor{magenta}{\textbf{Downstairs}}}
\hspace{1.2cm}
{\textcolor{purple}{\textbf{Upstairs}}}

    \begin{minipage}[b]{\textwidth}
        \centering
        \includegraphics[width=0.155\linewidth]{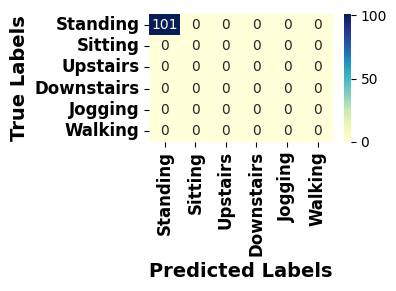}
        \includegraphics[width=0.155\linewidth]{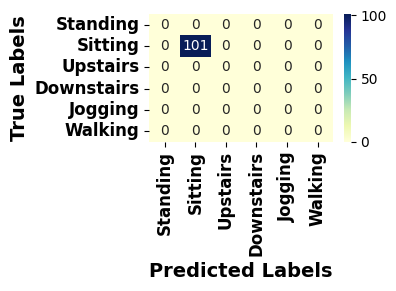}
        \includegraphics[width=0.155\linewidth]{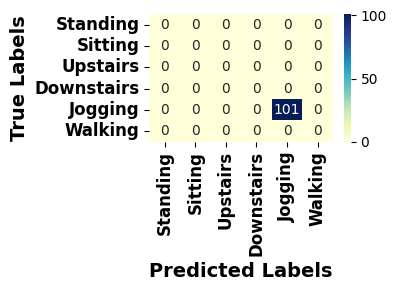}
        \includegraphics[width=0.155\linewidth]{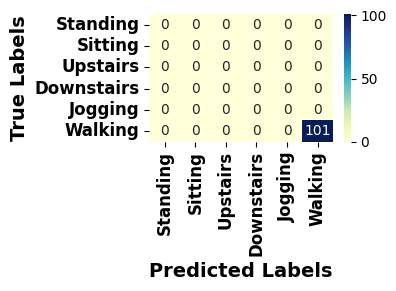}
        \includegraphics[width=0.155\linewidth]{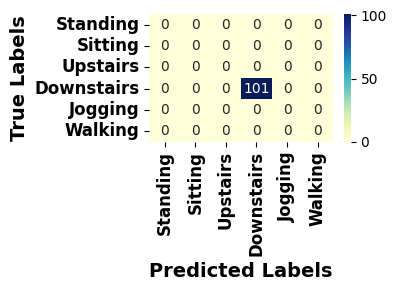}
        \includegraphics[width=0.155\linewidth]{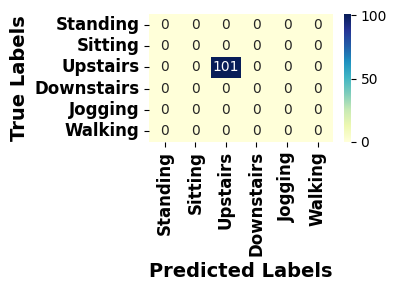} \\
    \end{minipage}
    \vspace{0.8em}

    \par\noindent
    \begin{minipage}[b]{\textwidth}
        \centering
        \includegraphics[width=0.15\linewidth]{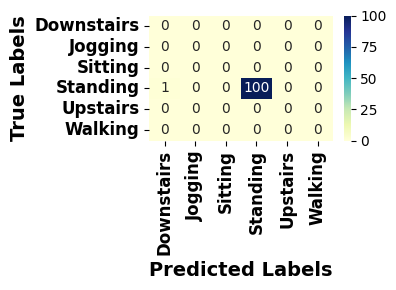}
        \includegraphics[width=0.15\linewidth]{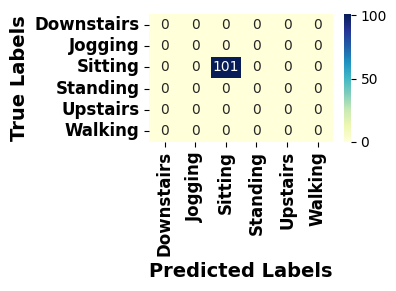}
        \includegraphics[width=0.15\linewidth]{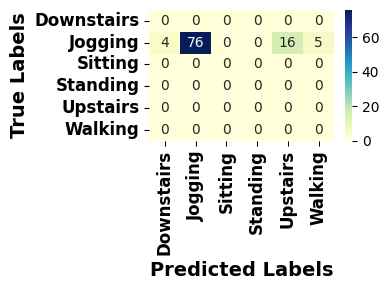}
        \includegraphics[width=0.15\linewidth]{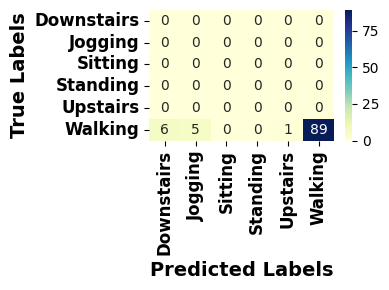}
        \includegraphics[width=0.15\linewidth]{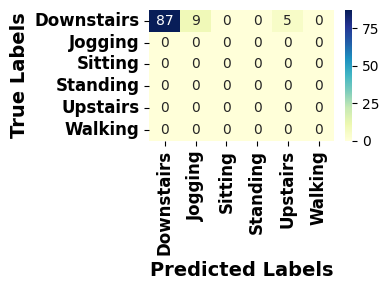}
        \includegraphics[width=0.15\linewidth]{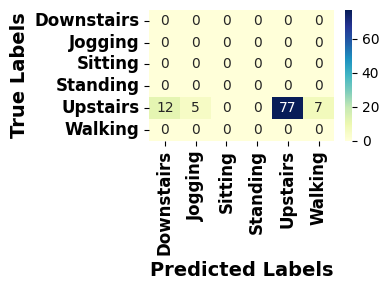} \\
    \end{minipage}
    \vspace{0.8em}

    \par\noindent
    \begin{minipage}[b]{\textwidth}
        \centering
        \includegraphics[width=0.15\linewidth]{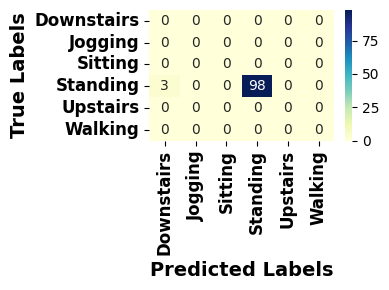}
        \includegraphics[width=0.15\linewidth]{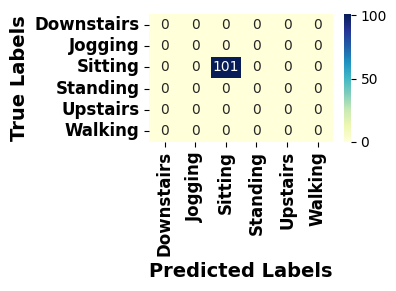}
        \includegraphics[width=0.15\linewidth]{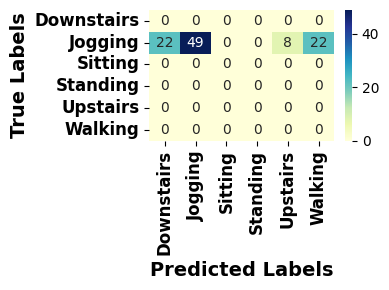}
        \includegraphics[width=0.15\linewidth]{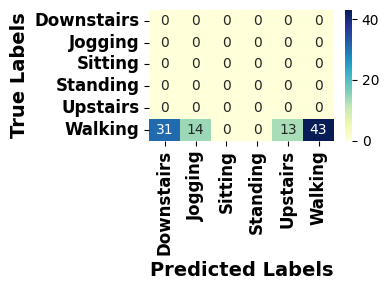}
        \includegraphics[width=0.15\linewidth]{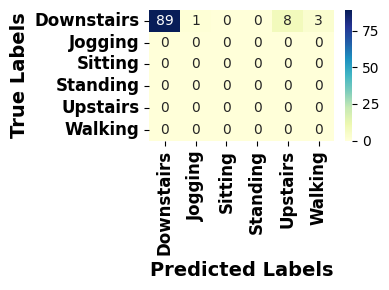}
        \includegraphics[width=0.15\linewidth]{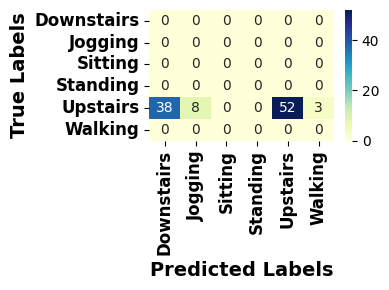} \\


    \end{minipage}
    \vspace{0.8em}

    \Description{Set of confusion matrices comparing human activity recognition performance across three models: KNN, LSTM, and TinyBERT on the MotionSense dataset. Each row corresponds to a model type, showing classification accuracy for six activities: Standing, Sitting, Jogging, Walking, Downstairs, and Upstairs. The color-coded labels visually distinguish activities and highlight differences in prediction accuracy across models.}
    \caption{Confusion matrices of Human Activities using KNN (top row) and LSTM (Middle row), and TinyBert (Bottom row) on the Motionsense
dataset. The activity labels are visually distinguished by colored boxes.}
    \label{fig:all-activity-models}
\end{figure*}

\begin{figure*}[h!]
    \centering

    \small
\hspace{0.5cm}
{\textcolor{blue}{\textbf{Standing}}}
\hspace{1.5cm}
{\textcolor{red}{\textbf{Sitting}}}
\hspace{1.3cm}
{\textcolor{cyan}{\textbf{Biking}}}
\hspace{1cm}
{\textcolor{orange}{\textbf{Walking}}}
\hspace{1cm}
{\textcolor{magenta}{\textbf{Stairsdown}}}
\hspace{1.2cm}
{\textcolor{purple}{\textbf{Stairsup}}}

    \begin{minipage}[b]{\textwidth}
        \centering
        \includegraphics[width=0.15\linewidth]{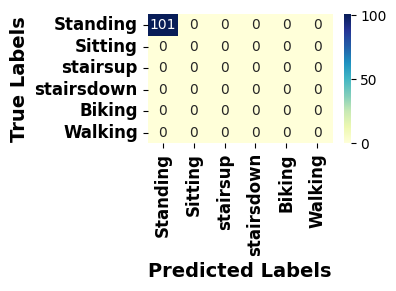}\includegraphics[width=0.15\linewidth]{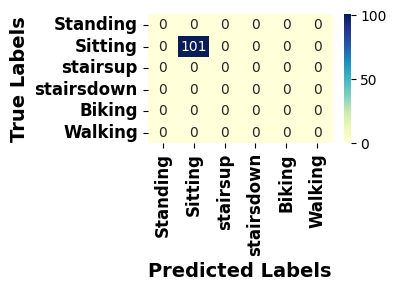}
        \includegraphics[width=0.15\linewidth]{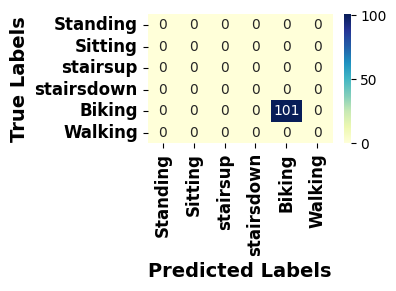}
        \includegraphics[width=0.15\linewidth]{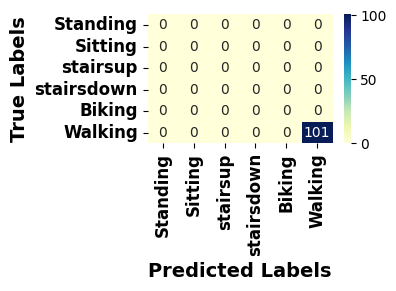}
        \includegraphics[width=0.15\linewidth]{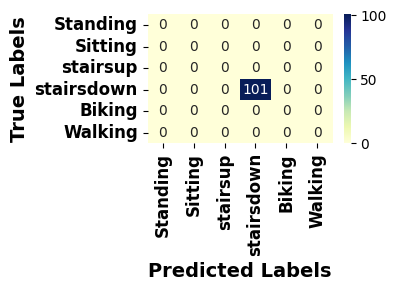}
        \includegraphics[width=0.15\linewidth]{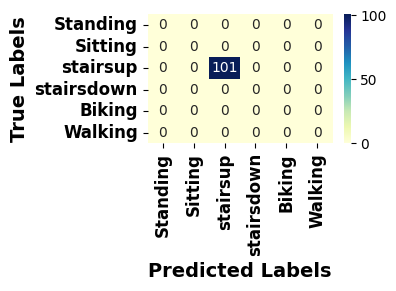} \\
    \end{minipage}
    \vspace{0.8em}

    \par\noindent
    \begin{minipage}[b]{\textwidth}
        \centering
        \includegraphics[width=0.15\linewidth]{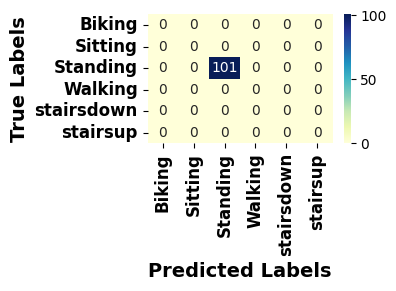}
        \includegraphics[width=0.15\linewidth]{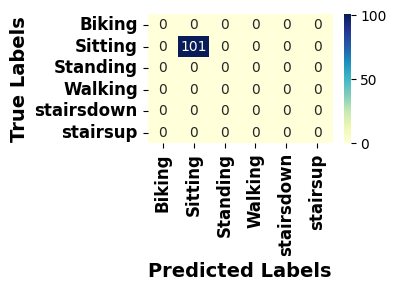}
        \includegraphics[width=0.15\linewidth]{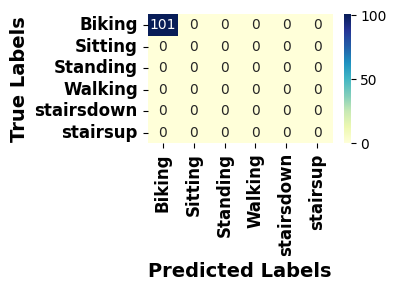}
        \includegraphics[width=0.15\linewidth]{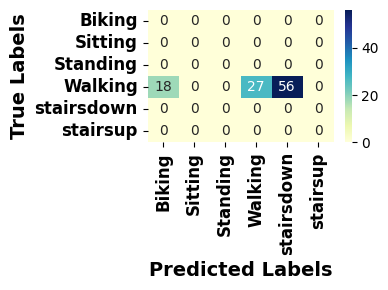}
        \includegraphics[width=0.15\linewidth]{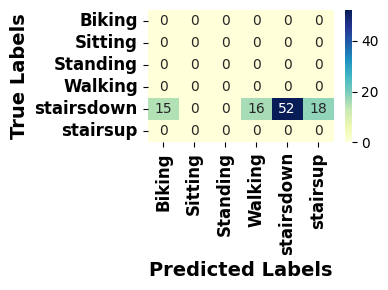}
        \includegraphics[width=0.15\linewidth]{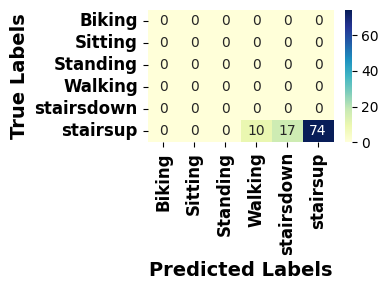} \\
    \end{minipage}
    \vspace{0.8em}

    \par\noindent
    \begin{minipage}[b]{\textwidth}
        \centering
        \includegraphics[width=0.15\linewidth]{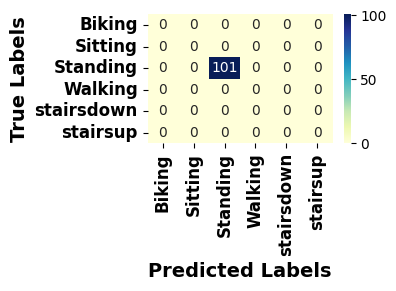}
        \includegraphics[width=0.15\linewidth]{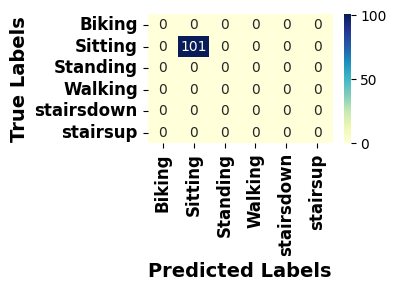}
        \includegraphics[width=0.15\linewidth]{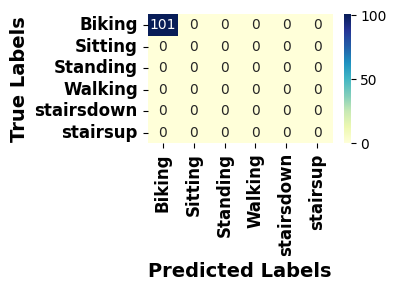}
        \includegraphics[width=0.15\linewidth]{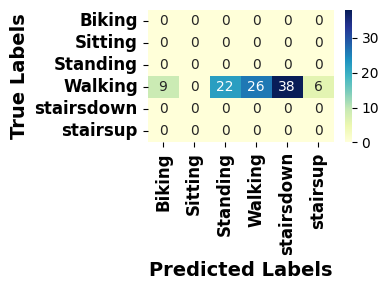}
        \includegraphics[width=0.15\linewidth]{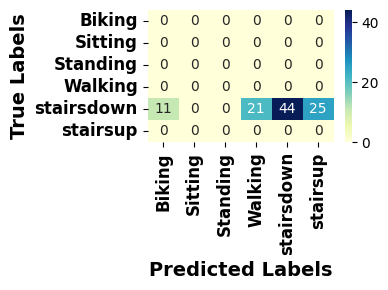}
        \includegraphics[width=0.15\linewidth]{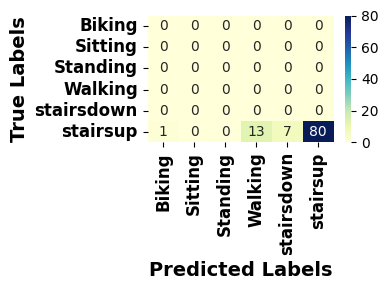} \\

    \end{minipage}
    \vspace{0.8em}
    \Description{Set of confusion matrices showing human activity recognition results using three models—KNN, LSTM, and TinyBERT—on the HHAR dataset. Each row corresponds to a model type, and each column represents one of six activities: Standing, Sitting, Biking, Walking, Stairsdown, and Stairsup. The figure visually compares classification accuracy across models, with color-coded labels highlighting activity categories.}
    \caption{ Confusion matrices of Human Activities using KNN (top row) and LSTM (Middle row), and TinyBert (Bottom row) on the HHAR dataset. The activity labels are visually distinguished by colored boxes.}
    \label{fig:all-activity-models-hhar}
\end{figure*}

\begin{figure*}[ht!]
    \centering
   \small
\hspace{0.5cm}
{\textcolor{blue}{\textbf{Standing}}}
\hspace{1.5cm}
{\textcolor{red}{\textbf{Sitting}}}
\hspace{1.3cm}
{\textcolor{cyan}{\textbf{Jogging}}}
\hspace{1.2cm}
{\textcolor{orange}{\textbf{Walking}}}
\hspace{1cm}
{\textcolor{magenta}{\textbf{Downstairs}}}
\hspace{1cm}
{\textcolor{purple}{\textbf{Upstairs}}}
    \begin{minipage}[b]{\textwidth}
        \centering
        \includegraphics[width=0.15\linewidth]{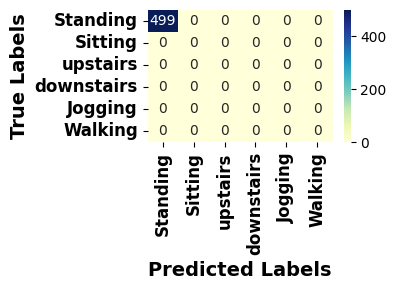}
        \includegraphics[width=0.15\linewidth]{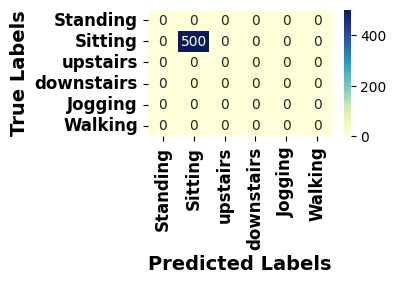}
        \includegraphics[width=0.15\linewidth]{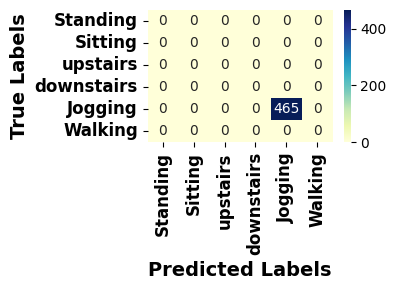}
        \includegraphics[width=0.15\linewidth]{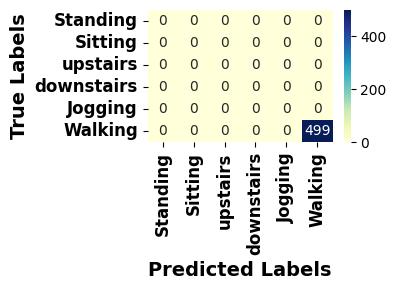}
        \includegraphics[width=0.15\linewidth]{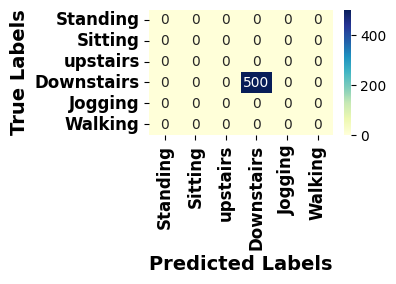}
        \includegraphics[width=0.15\linewidth]{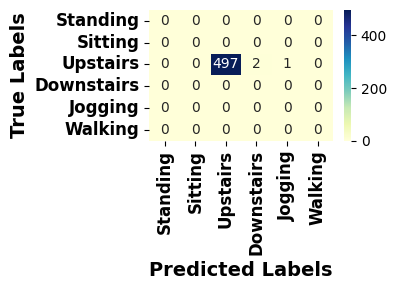} \\
    \end{minipage}
    \vspace{0.8em}

    \par\noindent
    \begin{minipage}[b]{\textwidth}
        \centering
        \includegraphics[width=0.15\linewidth]{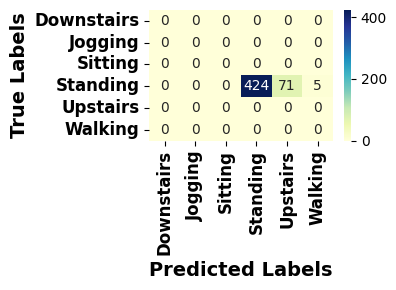}
        \includegraphics[width=0.15\linewidth]{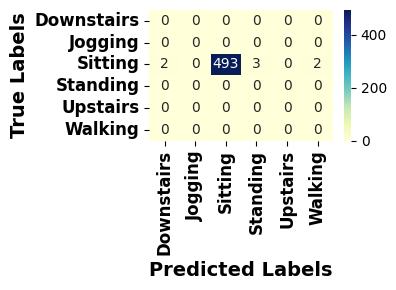}
        \includegraphics[width=0.15\linewidth]{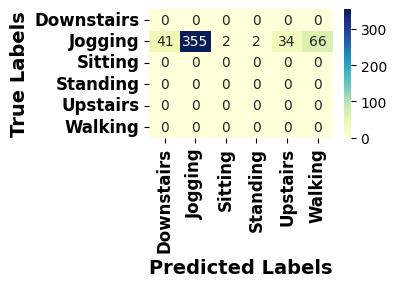}
        \includegraphics[width=0.15\linewidth]{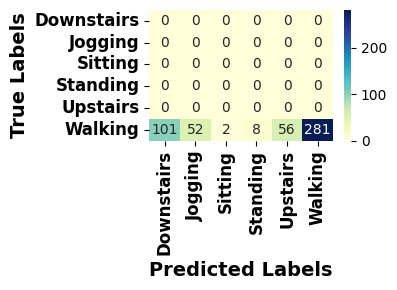}
        \includegraphics[width=0.15\linewidth]{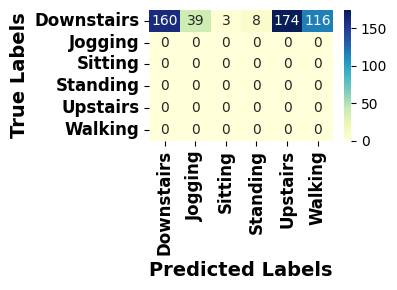}
        \includegraphics[width=0.15\linewidth]{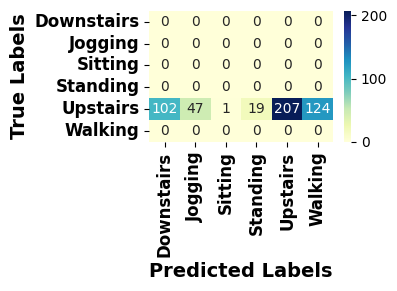} \\
    \end{minipage}
    \vspace{0.8em}

    \par\noindent
    \begin{minipage}[b]{\textwidth}
        \centering
        \includegraphics[width=0.15\linewidth]{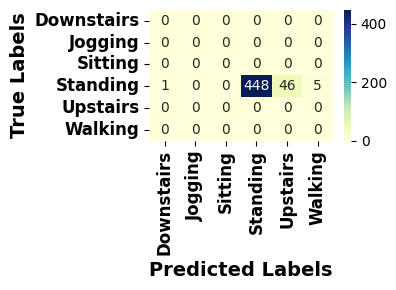}
        \includegraphics[width=0.15\linewidth]{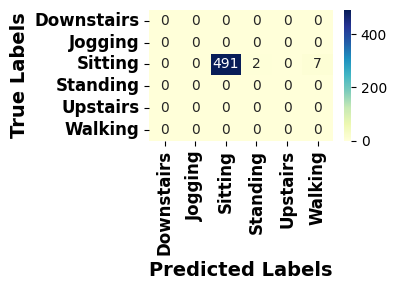}
        \includegraphics[width=0.15\linewidth]{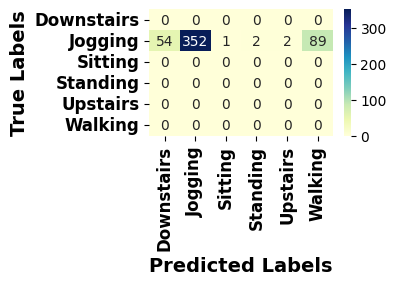}
        \includegraphics[width=0.15\linewidth]{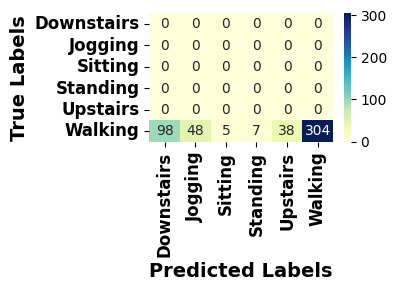}
        \includegraphics[width=0.15\linewidth]{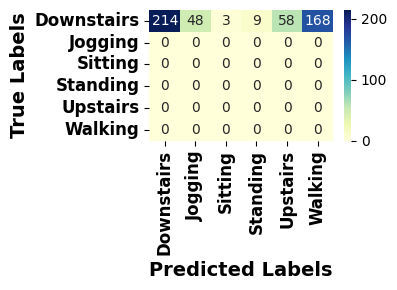}
        \includegraphics[width=0.15\linewidth]{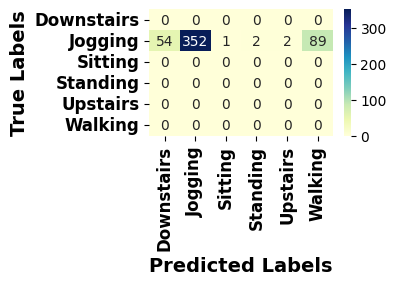} \\
        
    \end{minipage}
    \vspace{0.8em}
    \Description{Set of confusion matrices comparing the performance of KNN, LSTM, and TinyBERT models on the WISDM dataset. Each row represents a model type, and each column corresponds to one of six human activities: Standing, Sitting, Jogging, Walking, Downstairs, and Upstairs. The matrices visualize classification accuracy, with color-coded labels distinguishing activities and showing differences in model performance across motion patterns.}
    \caption{Confusion matrices of Human Activities using KNN (top row) and LSTM (Middle row), and TinyBert (Bottom row) on the WISDM dataset. The activity labels are visually distinguished by colored boxes.}
    \label{fig:all-activity-models-wisdm}
\end{figure*}

As shown in Table \ref{tab:transposed_accuracy_recall} and Fig. \ref{fig:all-activity-models}, \ref{fig:all-activity-models-hhar} and \ref{fig:all-activity-models-wisdm}, the KNN model consistently achieves high accuracy in identifying all activities across both the MotionSense and HHAR datasets, with only a slight decrease (0.99 probability) for the `Walking’ activity in the WISDM dataset. In comparison, the LSTM model demonstrates strong performance for specific activities, such as `Standing’, `Sitting’, and `Upstairs’ in the MotionSense dataset, but is less effective in sanitizing labels for `Downstairs’, `Walking’, and `Jogging’. This trend is also observed in the HHAR dataset, where the LSTM accurately classifies `Standing’, `Sitting’, and `Biking’, but has difficulty with `Stairsdown’, `Stairsup’, and `Walking’. For the WISDM dataset, the LSTM correctly identifies `Standing’, `Sitting’, `Walking’, and `Jogging’, yet struggles with `Upstairs’ and `Downstairs’. TinyBERT is achieves reliable detection for `Standing’ and `Sitting’ in the HHAR dataset but faces challenges in classifying activities such as `Upstairs’ and `Walking’. Overall, the KNN model demonstrates superior performance compared to both the LSTM and TinyBERT models across all evaluated datasets.

\begin{table}[ht]
\centering
\caption{\centering{Performance Comparison of LLM-Based and Traditional Methods on MotionSense, HHAR, and WISDM Datasets.}}
\label{tab:comparison_for_all}
\scriptsize

\begin{tabular}{p{1.25cm}|>{\centering}p{1cm}>{\centering}p{1cm}>{\centering}p{1cm}|>{\centering}p{1cm}>{\centering}p{1cm}>{\centering}p{1cm}|>{\centering}p{1cm}>{\centering}p{1cm}>{\centering\arraybackslash}p{1.5cm}}
\hline
\textbf{Model} & \multicolumn{3}{c|}{\textbf{MotionSense}} & \multicolumn{3}{c|}{\textbf{HHAR}} & \multicolumn{3}{c}{{\textbf{WISDM}}} \\ 
\cline{2-10}
 & \textbf{Sanit.Q.} & \textbf{Comm. Cost} & \textbf{Resp. Time} & \textbf{Sanit.Q.} & \textbf{Comm. Cost} & \textbf{Resp. Time} & \textbf{Sanit.Q.} & \textbf{Comm. Cost} & \textbf{Resp. Time} \\ 
\hline 
GPT-4-Zero      & \cellcolor{green!40}0.97    & \cellcolor{red!20}3121    & \cellcolor{red!20}25     & \cellcolor{green!30}0.83     & \cellcolor{red!20}3086    & \cellcolor{red!20}25    & \cellcolor{green!40}0.83      & \cellcolor{red!20}3100    & \cellcolor{red!20}24     \\

GPT-4-One       & \cellcolor{green!40}1.00     & \cellcolor{red!15}2879    & \cellcolor{red!15}25     & \cellcolor{green!40}1.00        & \cellcolor{red!15}3063    & \cellcolor{red!10}23    & \cellcolor{green!40}1.00     & \cellcolor{red!15}2800    & \cellcolor{red!10}22     \\

GPT-4-Few       & \cellcolor{green!40}1.00     & \cellcolor{yellow!20}2622 & \cellcolor{yellow!20}21  & \cellcolor{green!40}1.00        & \cellcolor{yellow!40}1754 & \cellcolor{yellow!30}13 & \cellcolor{green!40}1.00     & \cellcolor{yellow!20}1700 & \cellcolor{yellow!20}12  \\

KNN             & \cellcolor{green!40}1.00    & \cellcolor{green!50}0.00     & \cellcolor{green!50}0.01 & \cellcolor{green!50}1.00       & \cellcolor{green!50}0.00     & \cellcolor{green!50}0.01 & \cellcolor{red!40}0.63    & \cellcolor{green!50}0.00     & \cellcolor{green!50}0.01 \\

LSTM            & \cellcolor{red!40}0.83    & \cellcolor{green!50}0.00     & \cellcolor{green!50}0.01 & \cellcolor{red!40}0.74       & \cellcolor{green!50}0.00     & \cellcolor{green!50}0.01 & \cellcolor{red!40}0.80    & \cellcolor{green!50}0.00     & \cellcolor{green!50}0.01 \\

{TinyBERT}        & \cellcolor{red!40}0.59    & \cellcolor{green!50}0.00     & \cellcolor{green!50}0.01 & \cellcolor{red!40}0.63       & \cellcolor{green!50}0.00     & \cellcolor{green!50}0.01 & \cellcolor{red!40}0.62    & \cellcolor{green!50}0.00    & \cellcolor{green!50}0.01 \\
\hline
\end{tabular}
\end{table}

\subsection{Comparison of LLMs with Traditional Methods}

As shown in Fig. \ref{fig:accuracy}, Fig. \ref{fig:recall}, and Table \ref{tab:comparison_accuracy_sq}, ChatGPT-4 demonstrates high accuracy and sanitization quality in few-shot learning, establishing it as the most effective model for detecting and sanitizing data poisoning attacks when compared to ChatGPT-3.5 and Gemini. We further compare ChatGPT-4 with traditional methods based on sanitization quality, communication cost, and response time, as presented in Table \ref{tab:comparison_for_all}.

When comparing LLMs and traditional methods in terms of sanitization quality, which, in our experiments, is measured by recall for traditional models, the results indicate that KNN achieves a perfect sanitization quality of 1 on the MotionSense dataset and the HHAR dataset. In contrast, the sanitization quality of the LSTM model is lower than the KNN model. TinyBERT also achieves lower accuracy and sanitization quality across all datasets. Among the LLMs, ChatGPT-4 consistently exhibits the highest sanitization quality across all datasets.

Traditional methods have zero communication costs, as they do not process text data and operate locally. Additionally, their response times are lower, since only the model testing response time is considered. In contrast, ChatGPT-4 incurs higher communication costs and response times across all datasets. Notably, few-shot learning achieves lower communication costs and response times on the MotionSense and HHAR datasets, even though it requires sample data because it generates shorter responses compared to zero-shot and one-shot learning. However, for the WISDM dataset, ChatGPT-4 exhibits higher communication costs for both one-shot and few-shot scenarios.

Overall, the results clearly show that ChatGPT-4 outperforms traditional methods across all datasets. Furthermore, LLMs such as ChatGPT-4 provide flexible interfaces for natural language queries and explanations, supporting transparent decision-making and ease of integration in real-world applications.

\section{Conclusion}\label{sec:conclusion}

Traditional AI-based solutions for dealing with poisoning attacks in Internet of Things (IoT) systems, including wearable, sensor-based Human Activity Recognition (HAR), are limited by their reliance on curated, labeled datasets and lack adaptability to evolving data environments. In this paper, we present our research on harnessing the adaptability of LLMs to address these two critical limitations. Our proposed framework harnesses the capabilities of zero-shot, one-shot, and few-shot learning, enabling effective poisoning detection and sanitization without extensive retraining or additional data requirements. Through comprehensive evaluation, we demonstrate the framework’s adaptability, accuracy, and efficiency in poison detection and sanitizing poisoned data. We critically analyze the system's performance in key areas, including communication efficiency, latency, and privacy in design. Our findings underscore the potential of LLMs as powerful, scalable solutions for securing wearable IoT systems in dynamic, data-driven environments.

Importantly, while our focus is on HAR, the diversity and complexity encompassed by the three datasets utilized in the experiment, MotionSense, HHAR, and WISDM, support the potential generalizability of our framework to broader IoT domains. These datasets collectively introduce considerable diversity in user demographics, device types, and sensing environments. The MotionSense dataset includes inertial sensor data (accelerometer, gyroscope, gravity, and attitude) sampled at 50 Hz from 24 participants performing six activities in a controlled setting. Crucially, the participants vary widely in weight (48–102 kg), height (161–190 cm), age (18–46 years), and gender (13 male, 11 female), introducing intra-class variability that reflects user heterogeneity common in real-world IoT deployments. The HHAR dataset introduces substantial device diversity, with data collected from 4 smartwatches (2 LG watches, 2 Samsung Galaxy Gears) and 8 smartphones (2 Samsung Galaxy S3 mini, 2 Samsung Galaxy S3, 2 LG Nexus 4, 2 Samsung Galaxy S+), across 9 users performing six daily activities. These devices differ in sampling rates (typically 50–200Hz), orientations, and internal sensor calibrations, mimicking deployment challenges in heterogeneous IoT ecosystems. The WISDM dataset, gathered from 36 users, contains accelerometer data sampled at 20 Hz during semi-naturalistic usage of Android smartphones, with activities such as walking, jogging, and stair navigation. Together, the datasets span a wide range of environmental conditions (lab-controlled, semi-natural, and real-world), device types, and user demographics. This diversity positions our framework as a robust solution for addressing key challenges in real-world IoT deployments, such as sensor and user variability, device heterogeneity, and environmental complexity.

Future work will aim to refine the framework's performance in real-time applications, enhance privacy measures, and expand its application across diverse IoT ecosystems to solidify LLMs as a core component in the defense against data poisoning in wearable IoT. We will explore the trade-off between the performance of the LLM in dealing with different attacks on IoT systems and critical system performance metrics. Our proposed research makes several key contributions to the broader landscape of IoT. Specifically, it explores the potential of utilizing Large Language Models (LLMs) in the domain of IoT security, highlighting their role in enhancing the reliability and resilience of IoT-connected devices. Our work demonstrates the adaptability of LLMs in dynamic IoT environments while reducing the dependence on large, labeled datasets, making IoT security more efficient and scalable.

\bibliographystyle{ACM-Reference-Format}
\bibliography{references}

\end{document}